\documentclass[]{fairmeta}
\usepackage[disable]{todonotes}
\usepackage{wrapfig}
\usepackage{enumitem}
\usepackage{tabularx} 
\usepackage{listings}
\usepackage{multirow}
\usepackage{makecell}
\usepackage{graphicx}
\usepackage{wrapfig}
\usepackage{tabularx}
\usepackage{textcomp}
\usepackage{stfloats}
\usepackage{url}
\usepackage{verbatim}
\usepackage{titlesec}
\usepackage{tocloft}
\usepackage{adjustbox}
\usepackage{multirow}
\usepackage{pifont}
\usepackage{tikz}
\usepackage{comment}
\usepackage{amsmath,amssymb} %
\usepackage{colortbl}  %
\usepackage{color}
\usepackage{booktabs} 
\usepackage{hyperref}
\usepackage{graphicx}    %
\usepackage{subcaption} 
\usepackage{multirow} %
\usepackage{booktabs} %
\usepackage{subcaption} %
\RequirePackage{xspace}
\makeatletter
\DeclareRobustCommand\onedot{\futurelet\@let@token\@onedot}
\def\@onedot{\ifx\@let@token.\else.\null\fi\xspace}
\usepackage[most]{tcolorbox}
\usepackage{xcolor}
\usepackage{booktabs}
\usepackage{multirow}
\usepackage{makecell}
\usepackage{diagbox}
\usepackage{adjustbox}
\usepackage[table]{xcolor}
\usepackage{xltabular}
\setcitestyle{square,comma,numbers,sort&compress}
\usepackage{placeins}
\usepackage{booktabs}
\usepackage{makecell}
\usepackage{pifont}

\def\ie{\emph{i.e}\onedot}

\makeatother

\definecolor{adptorange}{RGB}{248, 205, 172}
\definecolor{cmpblue}{RGB}{189, 215, 238}
\definecolor{cmpblue}{RGB}{189, 215, 238}
\definecolor{rynn}{RGB}{108,92,186} %

\definecolor{our_red}{RGB}{232,157,160}
\definecolor{our_blue}{RGB}{136,206,230}
\definecolor{our_orange}{RGB}{246,200,168}
\definecolor{our_green}{RGB}{178,211,164}

\definecolor{attn_code0}{RGB}{247,215,200}
\definecolor{attn_code1}{RGB}{238,169,139}
\definecolor{mlp_code0}{RGB}{204,201,221}
\definecolor{mlp_code1}{RGB}{102,95,153}

\definecolor{token_blue}{RGB}{84, 120, 140}

\definecolor{myblue}{RGB}{233, 241, 249}
\definecolor{mygray}{RGB}{99, 110, 114}
\definecolor{myred}{RGB}{255, 118, 117}
\definecolor{myyellow}{RGB}{255, 234, 167}
\definecolor{mygreen}{RGB}{216, 226, 204}
\definecolor{mypurple}{RGB}{162, 155, 254}
\definecolor{mybrown}{RGB}{215, 190, 154}
\definecolor{myorange}{RGB}{255, 220, 190}

\usepackage{tabularx} %
\usepackage{pifont}       %
\usepackage{bbding}       %
\usepackage{fontawesome}
\usepackage{xspace}
\newcommand{\xmark}{\ding{55}}
\newcommand{\cmark}{\ding{51}}
\usepackage{float}

\newlength\savewidth

\newcolumntype{x}[1]{>{\centering\arraybackslash}p{#1pt}}
\newcolumntype{y}[1]{>{\raggedright\arraybackslash}p{#1pt}}
\newcolumntype{z}[1]{>{\raggedleft\arraybackslash}p{#1pt}}

\renewcommand{\paragraph}[1]{\vspace{1mm}\noindent\textbf{#1}}
\usepackage{colortbl}
\usepackage{xcolor}
\usepackage{wrapfig}
\usepackage{amssymb}

\renewcommand{\paragraph}[1]{\vspace{1.25mm}\noindent\textbf{#1}}

\usepackage{algorithm}
\usepackage{listings}

\definecolor{codeblue}{rgb}{0.25, 0.5, 0.5}
\definecolor{codekw}{rgb}{0.35, 0.35, 0.75}
\lstdefinestyle{Pytorch}{
    language = Python,
    backgroundcolor = \color{white},
    basicstyle = \fontsize{9pt}{8pt}\selectfont\ttfamily\bfseries,
    columns = fullflexible,
    aboveskip=1pt,
    belowskip=1pt,
    breaklines = true,
    captionpos = b,
    commentstyle = \color{codeblue},
    keywordstyle = \color{codekw},
}

\definecolor{green}{HTML}{009000}
\definecolor{red}{HTML}{ea4335}

\creflabelformat{equation}{#2#1#3}

\newcommand{\huggingface}{\raisebox{-1.5pt}{\includegraphics[height=1.05em]{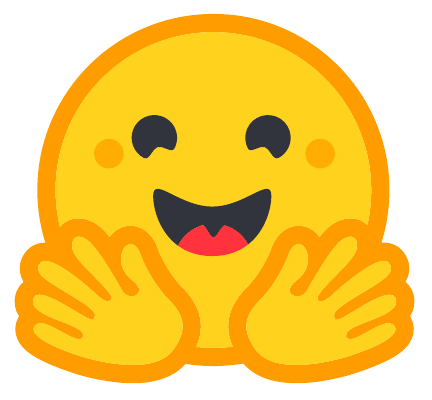}}\xspace}
\newcommand{\github}{\raisebox{-1.5pt}{\includegraphics[height=1.05em]{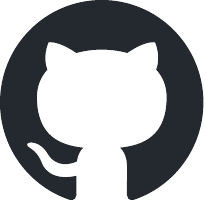}}\xspace}
\newcommand{\homepage}{\raisebox{-1.5pt}{\includegraphics[height=1.05em]{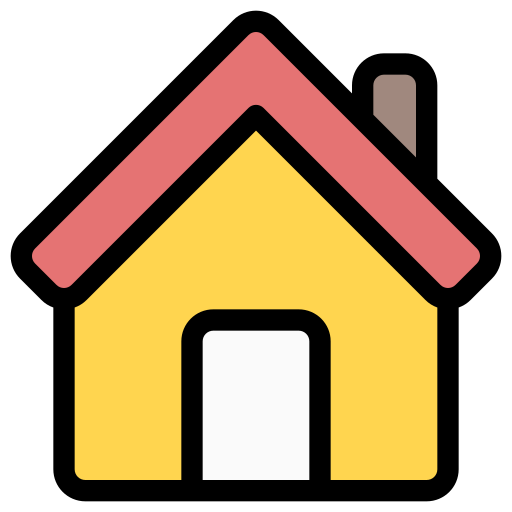}}\xspace}
\newcommand{\modelscope}{\raisebox{-1pt}{\includegraphics[height=0.95em]{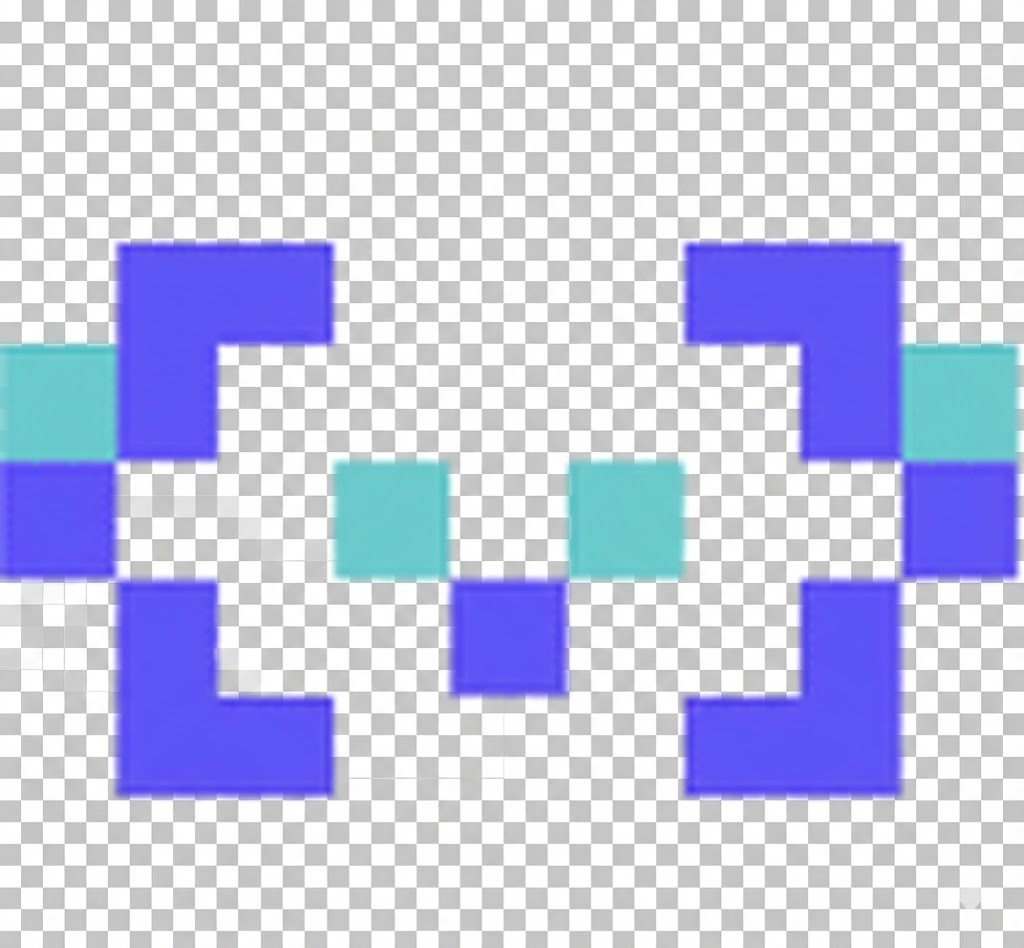}}\xspace}

\title{RynnValue: Scaling Robotic Value Foundation Models with Temporal Distance}

\author[*]{Dongchi Huang}
\author[*]{Hongyin Zhang}
\author[*]{Bohan Hou}
\author[\dagger]{Siteng Huang}
\author{Zhian Su}
\author{Hang Guo}
\author{Tong Lu}
\author{Zhaofeng Xu}
\author{Jiahao Tang}
\author{Jianfei Yang}
\author{Donglin Wang}
\author[\dagger]{Peixi Peng}
\author{Mingxiu Chen}
\author{Deli Zhao}
\author{Xin Li}

\affiliation[1]{DAMO Academy, Alibaba Group}
\affiliation[2]{Hupan Lab}
\contribution[*]{Equal contribution}
\contribution[\dagger]{Corresponding authors}

\abstract{

General-purpose reward models are increasingly the bottleneck for scaling robot learning, yet the recipe for learning value-related capabilities from large-scale heterogeneous corpora remains underexplored. Existing approaches tie supervision to task-internal anchors such as preferences or normalized progress, none of which transfer cleanly across embodiments and data sources. We introduce \textbf{RynnValue}, an open-source value foundation model for robotic manipulation that replaces these anchors with temporal distance, the directed cost-to-go from an observation to the language-specified goal. Because temporal-distance labels can be derived directly from timestamps, RynnValue scales to over 7,000 hours and roughly 3M instruction-conditioned clips without preference or progress annotations. To make temporal-value learning reliable at scale, we combine random temporal sampling, temporal-order shuffling, and value-isolation attention, suppressing shortcuts that would leave predictions insensitive to failures and regressions. Trained without preference labels, RynnValue attains an average Kendall's $\tau_a$ of 0.704 on RBM-EVAL-OOD, surpassing the fully preference-supervised state of the art (0.655) and more than doubling a progress-only counterpart (0.292), while generalizing zero-shot to unseen tasks, embodiments, and viewpoints.
As a zero-shot reward model, RynnValue serves a range of downstream applications. Converted into dense rewards via potential-based shaping, it raises real-world policy success from 52.5\% to 72.5\% online and from 63.8\% to 82.5\% offline; used for data filtering, it improves multi-task behavior cloning success from 35.0\% to 42.5\%; and applied as inference-time value guidance, it lifts a frozen policy's success from 67.5\% to 80.0\%.
These results establish temporal distance as a scalable supervision target and practical reward interface for generalist robot policies.

\begin{center}
    \renewcommand{\arraystretch}{1.2}
    \begin{tabular}{ll}
        \homepage  & \url{https://alibaba-damo-academy.github.io/RynnValue.github.io} \\
        \github  & \url{https://github.com/alibaba-damo-academy/RynnValue}\\
        \huggingface & \url{https://huggingface.co/collections/Alibaba-DAMO-Academy/rynnvalue} \\
        \modelscope  & \url{https://www.modelscope.cn/collections/DAMO_Academy/RynnValue}\\
    \end{tabular}
\end{center}

}

\date{\today}

\begin{document}
\thispagestyle{firstheader}
\maketitle
\pagestyle{empty}
\setcounter{tocdepth}{3}

\FloatBarrier
\vspace{-1em}

\FloatBarrier

\noindent

\begin{figure*}[ht]  %
    \centering
    \includegraphics[width=\textwidth]{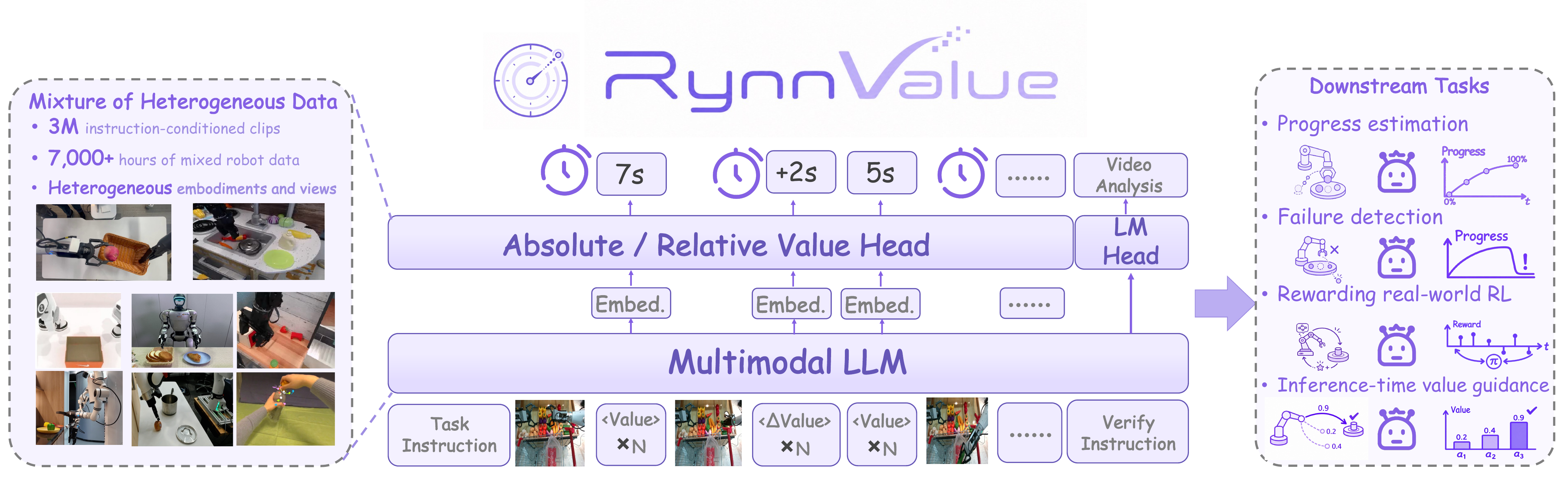}
    \caption{\textbf{Overview of RynnValue.}
RynnValue is a language-conditioned value model trained on over 7,000 hours of heterogeneous embodied data, comprising roughly 3M instruction-conditioned trajectory clips across diverse embodiments, viewpoints, and task domains. Given a language instruction and a sequence of sampled observations, the model builds an interleaved multimodal sequence of repeated absolute-value and relative-value queries, which RynnBrain encodes in a single forward pass. 
Two distributional heads then predict the absolute temporal distance to task completion and the signed relative temporal displacement between observations, while the language branch produces video analysis and language-conditioned verification. 
The resulting temporal values serve both as a unified interface for progress estimation, failure detection, and reward specification in reinforcement learning, and as a plug-and-play module for value guidance at inference time.
}
\label{fig:rynnvalue_arch}
\end{figure*}

\section{Introduction}
Generalist robot policies are increasingly trained with reinforcement learning~\cite{pi_0.5,intelligence2025pi}, yet scaling this loop is bottlenecked not by policy capacity but by reward supervision. Hand-designed rewards hardly generalize to open-ended tasks, while sparse success signals offer limited guidance for long-horizon behavior~\cite{yang2026rarm}. Existing general-purpose reward models rely on task-internal anchors such as preferences, reference demonstrations, or local state comparisons~\cite{yang2026rarm,zhanggeneralist,timerewarder2025learning,robodopamine2025general}, which tie supervision to particular trajectories or comparison sets and are difficult to reuse across heterogeneous data. The common fallback, normalized [0,1] progress~\cite{robometer2026scaling,roboreward2026general}, is an intra-trajectory coordinate rather than a goal-conditioned cost-to-go~\cite{schaul2015universal,puterman2014markov}, making it poorly aligned with the standard notion of value in control and difficult to maintain consistently across varying durations, embodiments, and task structures.

We argue for a shift in framing: from a reward model scoring trajectory-level anchors to a \textbf{value foundation model} predicting goal-conditioned cost-to-go as a single reusable interface. We instantiate this view in \textbf{RynnValue}, which adopts \emph{temporal distance} as its scaling target. Rather than normalizing progress within a trajectory, RynnValue estimates the directed temporal cost from the current observation to the language-specified goal. Under a minimum-time objective, this corresponds to the hitting-time cost-to-go, yielding clear directionality and task conditioning. It can further be converted into dense rewards through potential-based shaping~\cite{ng1999policy}, serving as a practical value interface rather than a task-specific heuristic.

Crucially, temporal distance scales naturally to heterogeneous data, since labels can be derived directly from timestamps once a completion cutoff is identified. Using this recipe, we construct a large training mixture of real-world, simulated, and egocentric trajectories spanning diverse embodiments, viewpoints, and task families. After preprocessing, the corpus exceeds 7,000 hours and comprises approximately 3M instruction-conditioned clips. Raw episode boundaries often do not reflect semantic task completion; we therefore apply subtask segmentation and cutoff relabeling to define coherent completion points. Unlike progress and comparison-based supervision~\cite{robometer2026scaling}, this recipe requires only instructions, timestamps, and relabeled cutoffs, without constructing preference pairs or progress annotations.

Scaling data alone, however, does not guarantee reliable temporal-value learning. In a multi-frame setting, the model may exploit regular sampling intervals, the presentation order of observations, or other value-query representations, rather than grounding predictions in task-relevant visual evidence. Such shortcuts can leave learned values insensitive to regressions, failures, and other non-monotonic events~\cite{geirhos2020shortcut}. RynnValue therefore incorporates complementary designs at two levels. At the visual level, we sample observations at irregular timestamps and perturb their temporal order during training, breaking the correspondence between sequence position and temporal distance. At the value level, we introduce value-isolation attention, so that each value-query group attends only to its own language-visual context and cannot access the value queries of other observations. Together, these designs suppress both temporal-order and value-extrapolation shortcuts, forcing each prediction to rely on the corresponding visual observation and task semantics. RynnValue further jointly supervises natural-language video description, instruction matching, and success prediction, adding sequence-level semantic cues alongside continuous temporal-value learning.

We evaluate RynnValue both as a standalone value model and as a reward interface for downstream policy learning. Trained without preference labels, RynnValue-8B attains an average Kendall's $\tau_a$ of 0.704 on RBM-EVAL-OOD, surpassing the fully preference-supervised state of the art (0.655) and more than doubling a progress-only counterpart (0.292), while generalizing zero-shot to unseen tasks, embodiments, and viewpoints. Converted into dense rewards via potential-based shaping, it raises real-world policy success from 52.5\% to 72.5\% online and from 63.8\% to 82.5\% offline over the strongest baseline. Moreover, RynnValue offers two scalable, plug-and-play routes to stronger real-world policies: filtering low-quality segments from demonstrations raises multi-task behavior-cloning success from 35.0\% to 42.5\%, while reranking candidate action chunks lifts a frozen policy's success from 67.5\% to 80.0\% without any policy or value-model updates.

In summary, our work makes the following contributions:

\begin{itemize}[labelsep=0.6em, leftmargin=1.2em,itemindent=0em]
\item We identify normalized progress as a supervision bottleneck for general-purpose robotic reward modeling, and reframe the problem as learning \textbf{a value foundation model} whose scaling target is \textbf{temporal distance (cost-to-go)}, a scalable, preference-free value target that unifies heterogeneous data under a single interface.
\item We present \textbf{RynnValue}, a robotic value foundation model trained with a label-cheap data recipe: labels are derived directly from timestamps, aided by subtask segmentation and cutoff relabeling, over 7,000 hours and roughly 3M instruction-conditioned clips spanning diverse embodiments, viewpoints, and task families.
\item We introduce complementary shortcut-suppression designs, \textbf{random temporal sampling}, \textbf{temporal-order shuffling},  and \textbf{value-isolation attention}, that keep temporal-value predictions grounded in visual evidence rather than spurious correlations, and pair them with dual distributional heads for absolute and relative temporal values.
\item We show that RynnValue \textbf{surpasses preference-supervised state-of-the-art} on both out-of-distribution trajectory ranking and instruction-trajectory alignment without any preference annotation, and serves as a practical value interface for downstream policy learning, improving real-world reinforcement learning via potential-based shaping, multi-task behavior cloning via data filtering, and frozen policies via inference-time value guidance.
\end{itemize}

\section{Related Work}
\label{sec:related_work}

\paragraph{Reward learning for robotic manipulation.}
Reward learning infers task objectives from demonstrations, preferences, or visual and temporal patterns, reducing the need for manual reward design. Inverse reinforcement learning recovers rewards from expert behavior~\cite{ng2000algorithms,abbeel2004apprenticeship,ziebart2008maximum} through joint cost-and-policy learning~\cite{finn2016guided} or adversarial distribution matching~\cite{ho2016generative,fu2018learning}, whereas preference-based methods use trajectory comparisons or ranked demonstrations~\cite{christiano2017deep,brown2019extrapolating}. Visual approaches support reward construction through temporal correspondence~\cite{sermanet2018time}, video representations~\cite{nair2022r3m,ma2023vip}, and language--image alignment~\cite{ma2023liv}. Moving from representations to reward signals, video-based methods derive rewards from human videos~\cite{chen2021learning}, progress ranking~\cite{yang2024rank2reward}, language-guided transfer~\cite{zhang2025rewind}, frame-wise temporal distance~\cite{timerewarder2025learning}, or video-language contrastive learning~\cite{sontakke2023roboclip}. Despite these advances, their applicability to diverse robotic scenarios remains a challenge. RynnValue aims to address this gap by learning a temporal value function that can generalize across tasks, embodiments, and viewpoints, providing a more versatile and reusable reward signal for robotic manipulation.

\paragraph{Robotic value foundation models.}
Recent efforts aim to provide reusable reward and value signals across robotic tasks, using trajectory-level comparisons or step-level process supervision. RoboReward~\cite{roboreward2026general} learns trajectory scores from annotated robotic corpora, while Robometer~\cite{robometer2026scaling} learns progress through trajectory comparisons. Step-level methods provide finer-grained feedback via normalized progress regression~\cite{robodopamine2025general,yang2026rarm,zhanggeneralist,ma2024generative} or advantage reward modeling~\cite{mao2026arm}. TOPReward~\cite{chen2026topreward} instead extracts progress rewards from pretrained vision-language models' task-completion token probabilities without reward-specific training. A central challenge is to achieve both scalability and strong performance: RoboReward and Robometer perform strongly but require large-scale annotations, whereas self-supervised methods derive progress targets from within-trajectory timestamps but remain limited on unseen tasks. RynnValue addresses this tradeoff by using temporal distance as a scalable, self-supervised target across heterogeneous trajectories. It further identifies and mitigates temporal-value shortcuts through a training recipe designed for robust value learning at scale.

\paragraph{Temporal distance learning.}
Temporal distance measures the time or transitions needed to reach a goal, giving goal-conditioned values a temporal interpretation. Goal-conditioned reinforcement learning~\cite{kaelbling1993goals} supports this formulation through universal value function approximators~\cite{schaul2015universal} and hindsight goal relabeling~\cite{andrychowicz2017hindsight}. Temporal-difference models learn distances to future observations~\cite{pong2018temporal}, and learned distances support planning via replay-buffer graph search~\cite{eysenbach2019search}. Transferable value representations include successor representations~\cite{dayan1993successor}, successor features with generalized policy improvement~\cite{barreto2017successor}, and universal successor features approximators~\cite{borsa2019usfa}. Contrastive learning connects representation inner products to goal-conditioned values~\cite{eysenbach2022contrastive}, quasimetric reinforcement learning captures asymmetric goal-reaching costs~\cite{wang2023quasimetric}, and contrastive successor features structure temporal distances for trajectory stitching~\cite{myers2024temporaldistance}. Translating them into a general-purpose robotic value foundation model requires handling heterogeneous robotic data and grounding temporal predictions in task semantics. RynnValue bridges this gap, providing the first robotic value foundation model that learns from temporal distance.

\section{Method}

As shown in \Cref{fig:rynnvalue_arch}, RynnValue is a language-conditioned value model that estimates the temporal distance from robot observations to task completion. RynnValue is built on RynnBrain~\cite{dang2026rynnbrain} and further pretrained on large-scale robot data, which provides representations adapted to embodied scenes, robot states, and language-conditioned manipulation tasks.

\subsection{RynnValue Dataset}
\label{sec:rynnvalue_dataset}

RynnValue unifies heterogeneous trajectories into state-level temporal-distance supervision. Learning temporal distance, however, requires more than a large collection of robot videos: each trajectory segment must be associated with a meaningful language goal and must describe task-relevant evolution toward that goal. We therefore construct the training corpus in three stages: assembling a heterogeneous data mixture (\Cref{sec:dataset_mixture}), converting raw episodes into instruction-conditioned segments with timestamp-derived labels (\Cref{sec:dataset_preprocessing}), and applying source-aware curation to remove invalid or task-irrelevant annotations (detailed in \Cref{sec:dataset_details}).

\subsubsection{Heterogeneous Data Mixture}
\label{sec:dataset_mixture}

\begin{table}[t]
\centering
\caption{
Composition of the heterogeneous data mixture. \# Episodes counts the original trajectories before subtask expansion, while \# Segments counts the instruction-conditioned trajectory segments obtained after subtask segmentation and cutoff relabeling. Overall, 1.67M original episodes are expanded into 3.09M trajectory segments. For sources without subtask annotations (segmentation source: coarse task), no splitting is applied and each episode is kept as a single segment.
}
\small
\setlength{\tabcolsep}{4pt}
\begin{tabular}{l r r r l}
\toprule
\textbf{Data Source} & \textbf{\# Episodes} & \textbf{\# Segments} & \textbf{\# Instructions} & \textbf{Segmentation Source} \\
\midrule
AgiBot~\cite{bu2025agibot_iros} & 167,535 & 1,166,042 & 3,741 & sub task \\
EgoDex~\cite{hoque2025egodex} & 338,234 & 338,234 & 2,038 & coarse task \\
Galaxea Open-World~\cite{jiang2025galaxea} & 16,979 & 95,671 & 11,070 & sub task \\
InternData-A1~\cite{tian2026interndata} & 320,905 & 320,905 & 348 & coarse task \\
Open X-Embodiment~\cite{open_x_embodiment_rt_x_2023} & 693,037 & 693,037 & 180,090 & coarse task \\
RDT~\cite{liu2025rdt} & 6,109 & 6,109 & 272 & coarse task \\
RoboCOIN~\cite{wu2025robocoin} & 67,420 & 410,877 & 2,124 & sub task \\
RoboMIND~\cite{wu2024robomind} & 32,138 & 32,138 & 184 & coarse task \\
RoboTwin~\cite{mu2025robotwin} & 27,414 & 27,414 & 23,527 & coarse \\
Soft-FOLD~\cite{zheng2025xvla} & 1,542 & 1,542 & 1 & coarse task \\
\midrule
\textbf{Total} & \textbf{1,671,313} & \textbf{3,091,969} & \textbf{223,395} & -- \\
\bottomrule
\end{tabular}
\label{tab:data_mixture}
\end{table}

To scale temporal-distance learning beyond curated task-specific demonstrations, we train RynnValue on a large heterogeneous mixture of robot, simulation, and egocentric data~\cite{bu2025agibot_iros,hoque2025egodex,jiang2025galaxea,open_x_embodiment_rt_x_2023}. The mixture exceeds 7,000 hours of trajectories and spans a broad range of manipulation settings, including single-arm platforms, dual-arm mobile manipulators, bimanual tabletop robots, dexterous-hand systems, simulated embodiments, and first-person human demonstrations. It also covers diverse camera configurations. This diversity exposes the model to large variations in morphology, viewpoint, control frequency, execution speed, task duration, object distribution, and language annotation granularity. \Cref{tab:data_mixture} summarizes the composition of the mixture.

Before subtask expansion, the corpus contains 1.67M original episodes. After subtask segmentation and cutoff-based relabeling, these data are further converted into over 3M instruction-conditioned trajectory segments, each associated with a language goal and a temporally localized completion target. Crucially, these sources are not unified by manually defining dataset-specific progress scales. Instead, they are mapped onto the same temporal-distance supervision interface: each trajectory segment provides timestamp-derived state-level labels measuring the remaining temporal distance to its completion cutoff.

\subsubsection{Data Preprocessing and Temporal-Distance Relabeling}
\label{sec:dataset_preprocessing}

Large-scale robot datasets differ substantially in how episodes and subtasks are defined~\cite{bu2025agibot_iros,open_x_embodiment_rt_x_2023,cadene2024lerobot}. We therefore convert each data source into instruction-conditioned trajectory segments. Long demonstrations are split using native temporal annotations when available; otherwise, the complete episode is retained as a coarse segment. Because recorded trajectories may contain post-completion motions, we further assign each segment a completion cutoff. The segment endpoint is used by default, while dataset-specific ratio- or duration-based trimming is applied when necessary to better approximate the first semantically completed observation~\cite{robometer2026scaling}.

Temporal-distance labels are then generated directly from timestamps: observations before the cutoff are labeled by their temporal distance to completion, while observations at or after the cutoff receive zero. This provides unified, dense supervision without dataset-specific progress normalization.
Separately, we use Qwen3-VL-27B~\cite{qwen3vl} to generate a segment-level description of the observed behavior and task-relevant events. These captions supervise the natural-language video-analysis output (\ie, \texttt{Video Description} in \Cref{eq:language_outputs}) and do not affect the temporal targets.

\subsection{Model Architecture}
\label{sec:model_architecture}
\paragraph{Multi-granularity value estimation.}
Given a text prompt $\ell$ comprising the task instruction and system prompts
for absolute and relative temporal estimation, embodiment metadata $m$, and
a sequence of $K$ observations $\mathcal{I}=\{I_{t_i}\}_{i=1}^{K}$,
RynnValue jointly estimates the temporal distance to task completion for each
observation and the signed temporal displacement between adjacent observations
in the presented sequence.
For each observation $I_{t_i}$, we insert an absolute-value query group
$\mathbf{V}_i=[V_{i,1},\ldots,V_{i,N}]$ consisting of repeated
\texttt{\textless value\textgreater} tokens. For $i\geq 2$, we also insert
a relative-value query group $\mathbf{R}_{i-1}=[R_{i-1,1},\ldots,R_{i-1,N}]$
of repeated \texttt{\textless relative\_value\textgreater} tokens before
$\mathbf{V}_i$ to estimate the displacement from $I_{t_{i-1}}$ to $I_{t_i}$.
Following~\cite{hou2025fire}, we use $N$ repeated query tokens for each
temporal prediction to alleviate the information bottleneck of a single
token and support a richer temporal representation.
Our main setting uses $N=8$ tokens per query group and $K=8$ observations
sampled from each video.
After the final temporal-query group, we append a natural-language
verification prompt $\mathbf{p}_{\mathrm{ver}}$, yielding the input sequence:

\begin{equation}
\mathbf{x}
=
\big[
m,\ell,\,
I_{t_1},\mathbf{V}_1,\,
I_{t_2},\mathbf{R}_1,\mathbf{V}_2,\,
\ldots,\,
I_{t_K},\mathbf{R}_{K-1},\mathbf{V}_K,\,
\mathbf{p}_{\mathrm{ver}}
\big].
\label{eq:temporal_query_sequence}
\end{equation}

The verification prompt instructs the model to first analyze the video in
natural language, then determine whether the observed behavior matches the
task instruction and whether the task is successfully completed:
\begin{equation}
\mathbf y_{\mathrm{lang}}
=
\big[
\texttt{Video Description: }\mathbf y^{\mathrm{vid}},\,
\texttt{Match: }\mathbf y^{\mathrm{match}},\,
\texttt{Success: }\mathbf y^{\mathrm{succ}}
\big].
\label{eq:language_outputs}
\end{equation}
This autoregressive ordering encourages the model to describe robot behavior,
object interactions, and task-relevant events, including evidence of completion,
before producing the final task-matching and task-success judgments.

\paragraph{Distributional temporal readouts.}
The $N$ tokens in each temporal-query group attend bidirectionally. We
concatenate their hidden states to retain complementary information across positions:
\begin{equation}
\widetilde{\mathbf h}^{V}_{i}
=
H_{\theta}(\mathbf{x})_{\operatorname{pos}(V_{i,1})}
\Vert \cdots \Vert
H_{\theta}(\mathbf{x})_{\operatorname{pos}(V_{i,N})},
\qquad
\widetilde{\mathbf h}^{R}_{i}
=
H_{\theta}(\mathbf{x})_{\operatorname{pos}(R_{i,1})}
\Vert \cdots \Vert
H_{\theta}(\mathbf{x})_{\operatorname{pos}(R_{i,N})},
\label{eq:temporal_query_concatenation}
\end{equation}
where $H_{\theta}$ denotes RynnBrain contextual representations and $\Vert$
denotes feature concatenation. With per-token hidden size $d$, two heads map
the $Nd$-dimensional representations to the bin logits:
\begin{equation}
\mathbf z^{V}_{i}
=
\operatorname{ValueHead}
\left(\widetilde{\mathbf h}^{V}_{i}\right),
\qquad
\mathbf z^{R}_{i}
=
\operatorname{RelativeHead}
\left(\widetilde{\mathbf h}^{R}_{i}\right).
\label{eq:temporal_readouts}
\end{equation}
The heads estimate temporal distance to the relabeled completion cutoff and
signed temporal displacement between adjacent observations in the presented
sequence, providing global and local temporal supervision, respectively.

Absolute targets span $[0,636.3]$ seconds and concentrate near zero with a long
right tail. A fixed quantile support provides finer resolution near zero while
covering the tail: $256$ bin centers sample the inverse empirical cumulative
distribution function (CDF) of temporal targets at equally spaced
probability levels. Mirroring the non-negative quantile knots around zero
gives a symmetric relative support over $[-636.3,636.3]$ seconds.
Following Dreamerv3~\cite{hafner2025dreamerv3}, we encode continuous targets as
two-hot distributions over adjacent bin centers and train with cross-entropy,
avoiding gradients that scale directly with temporal error. Inference takes
the expected bin centers~\cite{farebrother2024stop}:
\begin{equation}
v_i
=
\sum_{b=1}^{|\mathcal B_V|}
c^{V}_{b}\,
\operatorname{softmax}(\mathbf z^{V}_{i})_b,
\qquad
\Delta_i
=
\sum_{b=1}^{|\mathcal B_R|}
c^{R}_{b}\,
\operatorname{softmax}(\mathbf z^{R}_{i})_b,
\label{eq:absolute_relative_decode}
\end{equation}
where $\mathcal B_V$ and $\mathcal B_R$ are the absolute and relative bin sets,
and $c^{V}_{b}$ and $c^{R}_{b}$ are their centers in seconds.

\begin{figure*}[t]
    \centering
    \includegraphics[width=\textwidth]{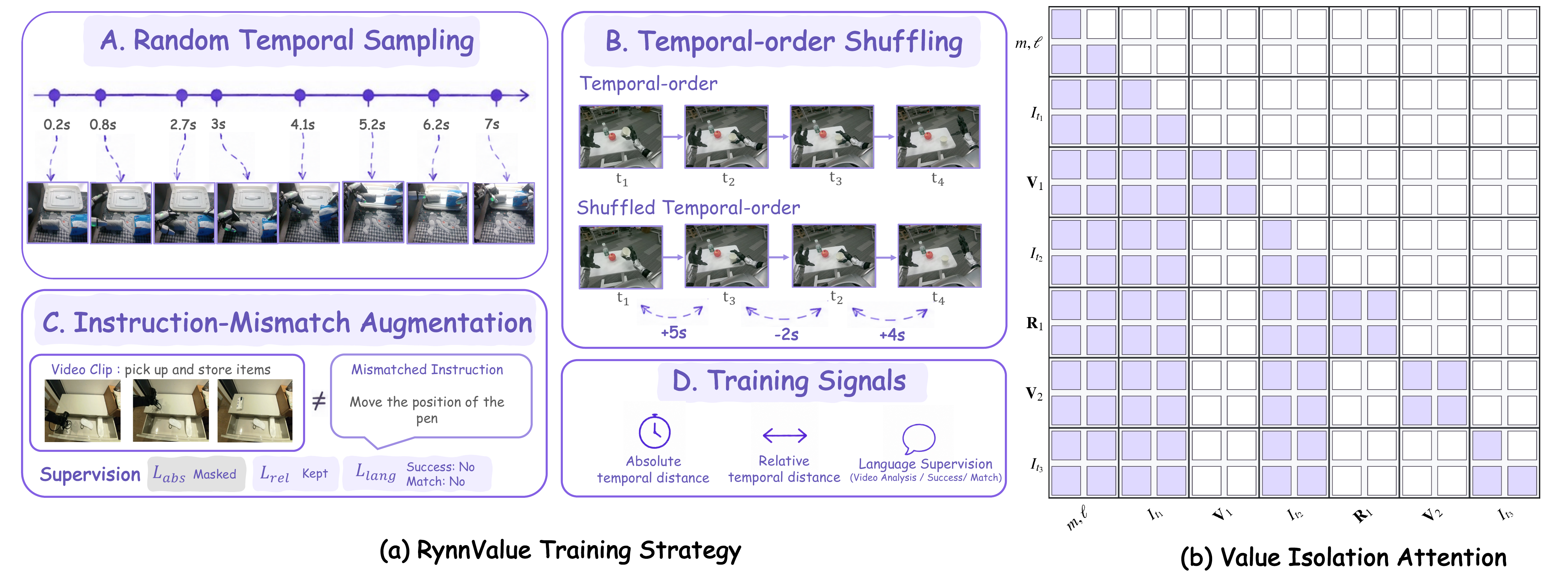}
    \caption{
    RynnValue training pipeline and value-isolation attention. \textbf{(a)} Training strategy. Random temporal sampling and temporal-order shuffling suppress shortcuts tied to sampling intervals and sequence position, while instruction-mismatch augmentation strengthens language–visual grounding. RynnValue jointly learns absolute temporal distance, relative temporal displacement, and natural-language supervision. \textbf{(b)} Value-isolation attention. Within each value-query group, repeated queries attend to one another and to the language–visual context, while remaining isolated from other value-query groups. Colored cells denote visible attention connections.
}
    \label{fig:rynnvalue_training}
\end{figure*}

\subsection{Shortcut learning}
\label{sec:sampling_strategy}

Scaling data alone does not guarantee reliable temporal-value learning, because a multi-frame model can exploit regularities in the input rather than the visual evidence. We combine random temporal sampling, temporal-order shuffling and value-isolated attention to suppress shortcuts tied to temporal intervals, sequence positions, and value extrapolation.

\paragraph{Random temporal sampling.}
For each instruction-conditioned trajectory clip, we randomly sample $K=8$
observations at irregular timestamps. The resulting non-uniform temporal gaps
break the near-arithmetic value patterns induced by uniform sampling and
discourage the model from exploiting fixed sampling intervals as a shortcut.

\paragraph{Temporal-order shuffling.}
As illustrated in~\Cref{fig:rynnvalue_training}(a), we shuffle the chronological order of sampled observations. Half of the training sequences are independently sampled without temporal sorting, while the remainder follow a forward-biased temporal walk with occasional backward transitions, whose per-step probability we refer to as the rewind probability. Relative temporal targets are computed between adjacent observations in the presented order and can therefore be either positive or negative. Shuffling prevents the model from regressing a stereotypical value curve from sequence positions.

\paragraph{Value-isolation attention.}
Without an explicit attention constraint, a temporal query can predict its value by extrapolating from previously exposed value tokens rather than by interpreting the corresponding visual evidence. This shortcut produces smooth value curves while leaving predictions insensitive to regressions, failures, and other non-monotonic events. Inspired by attention masking strategies that explicitly control information flow and prevent target leakage~\cite{yang2019xlnet,dong2019unified}, we introduce value-isolation attention, illustrated in~\Cref{fig:rynnvalue_training}(b). Absolute and relative temporal queries serve as pooling tokens that aggregate information from the preceding language--visual context. Within each temporal-query group, tokens attend to one another to integrate complementary features before concatenation, while remaining isolated from queries in all other groups. Subsequent observation tokens are also prevented from attending to temporal queries, blocking indirect information leakage to later temporal-query groups through the observation representations.

\subsection{Instruction-mismatch augmentation.}
A value model should also recognize when the language instruction does not describe the observed video, rather than always reporting smooth progress toward the stated goal.
As shown in~\Cref{fig:rynnvalue_training}, we therefore introduce instruction-mismatch augmentation. For 10\% of the training samples, we replace the original instruction with one sampled from a different trajectory and supervise the language branch to predict \texttt{Match: No} and \texttt{Success: No}. Because the original completion cutoff is no longer valid under the substituted instruction, we mask the absolute temporal-distance loss (\Cref{eq:absolute_training_loss}) to avoid introducing incorrect supervision.
The relative temporal loss (\Cref{eq:relative_training_loss}) is retained, as the relative target measures the temporal displacement between observations and is independent of the task instruction.
This augmentation teaches RynnValue to detect instruction-video mismatches and remain reliable under substantial language-visual inconsistency.

\subsection{Joint Training Objectives}

RynnValue is jointly optimized using three cross-entropy objectives: an
absolute temporal-distance loss, a relative temporal-distance loss, and a
causal language-modeling loss. Although all three objectives use
cross-entropy, the first two operate over distributional temporal bins, whereas
the third operates over the language vocabulary.

\paragraph{Temporal targets.}
Let $t_G$ denote the relabeled completion cutoff and $t_i$ the timestamp of observation $I_{t_i}$. We define its absolute temporal-distance target as $v_i^\star=\max(0,t_G-t_i)$, and the relative target between two consecutively presented observations as $\Delta_i^\star=t_{i+1}-t_i$. Accordingly, $\Delta_i^\star>0$ indicates forward temporal progress, whereas $\Delta_i^\star<0$ indicates temporal regression. Here, ``consecutive'' refers to adjacency in the presented multimodal sequence, not in the original video.
Both continuous targets are encoded as two-hot distributions over $256$-bin quantile supports, denoted $\mathcal B_V$ and $\mathcal B_R$ for the absolute and relative supports, respectively.

\paragraph{Absolute temporal-distance loss.}
The absolute temporal-distance objective is
\begin{equation}
\mathcal L_{\mathrm{abs}}
=
-\frac{\omega}{K}
\sum_{i=1}^{K}
\sum_{b=1}^{|\mathcal B_V|}
\left[
\operatorname{TwoHot}_{\mathcal B_V}(v_i^\star)
\right]_b
\log
\left[
\operatorname{softmax}(\mathbf z_i^V)
\right]_b ,
\label{eq:absolute_training_loss}
\end{equation}
where $\mathbf z_i^V$ denotes the logits produced by the absolute-value head, and the decoded continuous prediction is $v_i$. The sample-level mask $\omega$ is set to $1$ for instruction-matched examples and $0$ for instruction-mismatched examples, for which the original completion cutoff is no longer valid. %

\paragraph{Relative temporal-distance loss.}
The relative temporal-distance objective is
\begin{equation}
\mathcal L_{\mathrm{rel}}
=
-\frac{1}{K-1}
\sum_{i=1}^{K-1}
\sum_{b=1}^{|\mathcal B_R|}
\left[
\operatorname{TwoHot}_{\mathcal B_R}(\Delta_i^\star)
\right]_b
\log
\left[
\operatorname{softmax}(\mathbf z_i^R)
\right]_b ,
\label{eq:relative_training_loss}
\end{equation}
where $\mathbf z_i^R$ denotes the logits produced by the relative-value head, and the corresponding decoded prediction is denoted by $\Delta_i$. The absolute objective anchors each observation to task completion, whereas the relative objective captures forward and backward temporal displacement between consecutively presented observations.

\paragraph{Natural-language objective.}
For natural-language supervision, RynnValue autoregressively predicts the structured output defined in~\Cref{eq:language_outputs} and is optimized using standard token-level cross-entropy loss:
\begin{equation}
    \mathcal L_{\mathrm{lang}}
    =
    -\frac{1}{|\mathcal Y|}
    \sum_{j\in\mathcal Y}
    \log
    p_{\theta}
    \left(
    y_j
    \mid
    \mathbf x,y_{<j}
    \right),
    \label{eq:language_training_loss}
\end{equation}
where $\mathcal Y$ denotes the supervised language-token positions. Success
prediction is therefore learned through the natural-language objective,
without introducing a separate success-classification head.

\paragraph{Joint objective.}
The final training objective combines the three cross-entropy losses:
\begin{equation}
    \mathcal L
    =
    \mathcal L_{\mathrm{abs}}
    +
    \mathcal L_{\mathrm{rel}}
    +
    \lambda_{\mathrm{lang}}
    \mathcal L_{\mathrm{lang}},
    \qquad
    \lambda_{\mathrm{lang}}=2.
    \label{eq:joint_training_objective}
\end{equation}
The absolute and relative losses update their corresponding distributional
heads and the visual-language backbone. 
Further dataset curation details and the construction of the frozen temporal-distance quantile supports are provided in \Cref{sec:dataset_details,sec:model_details}.

\subsection{Inference and Reward Interface}
\label{sec:inference_reward_interface}

\paragraph{Chronological inference.}
At inference time, the training-time sampling augmentations are disabled and observations are arranged in chronological order. We query the model with a standard causal mask rather than the value-isolation mask used during training, which lets us reuse the default inference kernels without any custom masking; \Cref{app:attention_ablation} audits both inference masks and shows that the causal mask also yields stronger trajectory-ranking scores. RynnValue then decodes its distributional outputs into continuous temporal-distance estimates, while the language branch produces the video description, instruction-video matching judgment, and success judgment. In trajectory-ranking evaluation, the matching judgment is used only to gate instruction-video mismatches; the remaining language outputs do not enter the ranking score.

\paragraph{Temporal-distance potential.}
RynnValue predicts temporal distance rather than reward directly. Let $v_t$
denote the final temporal-distance estimate for observation $I_t$,
after distributional decoding.
The original
prediction remains non-negative: a large $v_t$ indicates that more time is
required to complete the task, whereas $v_t=0$ corresponds to the predicted
completion boundary. To match the conventional value semantics that
\emph{larger is better}, we convert temporal distance into an observation
potential by reversing its sign:
\begin{equation}
    \Phi_t
    =
    \Phi_{\theta}(I_t,\ell,m)
    =
    -v_t,
    \label{eq:rv_potential}
\end{equation}
where $\ell$ and $m$ denote the task instruction and metadata, respectively.
Consequently, states before task completion have negative potential,
while the potential approaches zero as the agent reaches the goal. This
sign reversal preserves the temporal scale rather than normalizing predictions
to a task-specific $[0,1]$ interval.

\section{Benchmark Evaluation}
\label{sec:benchmark_eval}
We evaluate RynnValue on the RBM-EVAL-OOD test suite introduced by Robometer~\cite{robometer2026scaling}. The benchmark contains 976 trajectories from six out-of-distribution datasets spanning different institutions, robot embodiments, camera viewpoints, and task families. For each task, the benchmark provides trajectories annotated with different execution-quality levels, including failed, suboptimal, and successful executions. Performance is measured by Kendall's $\tau_a$ between the ground-truth trajectory-quality ordering and the ordering induced by the predicted scores.
We describe the baselines in \Cref{app:eval_baselines} and provide additional reward evaluation details in \Cref{sec:eval_details}.

\subsection{Experimental Setup}
\label{sec:experimental_setup}

\paragraph{Model configuration.}
In \Cref{tab:rbm_ood_policy_ranking} we additionally report a RynnValue-4B variant to probe how performance scales with backbone size; all other experiments use the 8B model.
Each training sample contains $K=8$
observations, with eight repeated query tokens for each absolute or relative
prediction slot. Both temporal heads are implemented as BroNet \cite{NEURIPS2024_cd3b5d2e} residual MLPs
with hidden width $4096$, depth $8$, and ReLU activations. The absolute and
relative temporal targets are represented using 256-bin quantile supports over
$[0,636.3]$ and $[-636.3,636.3]$ seconds, respectively; their construction and
empirical statistics are detailed in \Cref{app:temporal_quantile_supports}.

\paragraph{Optimization.}
We jointly optimize the absolute temporal-distance, relative temporal-distance,
and natural-language objectives as shown in~\Cref{eq:joint_training_objective}.
RynnValue is trained with AdamW using a learning rate of $1\times10^{-6}$,
$\beta_1=0.9$, $\beta_2=0.95$, weight decay $0.1$, and
$\epsilon=10^{-8}$. We use a constant learning-rate schedule without warm-up
and clip the gradient norm at $100$. Training is conducted in bfloat16 with
FSDP hybrid sharding and a per-device batch size of $2$.

During training, temporal-order shuffling is applied with probability $0.5$,
and the forward-biased sampling process uses a rewind probability of $0.3$.
Instruction-mismatch augmentation is applied to $10\%$ of the samples. For
mismatched samples, the absolute temporal-distance loss is masked, while the
relative and natural-language objectives are retained.

\subsection{Main Results}

\begin{table*}[t]
    \centering
    \caption{
        Per-dataset trajectory-ranking results on the RBM-EVAL-OOD test suite,
        measured by Kendall's $\tau_a$ ($\uparrow$).
        Bold values indicate the best overall results.
        $\dagger$ denotes the best result among methods trained without explicit
        trajectory-level preference supervision, i.e., progress/value-only methods.
        Baseline results are taken from Robometer~\cite{robometer2026scaling}.
    }
    \label{tab:rbm_ood_policy_ranking}
    \small
    \setlength{\tabcolsep}{4.5pt}
    \renewcommand{\arraystretch}{1.12}
    \resizebox{\textwidth}{!}{
    \begin{tabular}{lccccccc}
        \toprule
        \textbf{Method}
        & \textbf{USC Franka}
        & \textbf{USC Koch}
        & \textbf{USC Trossen}
        & \textbf{USC xArm}
        & \textbf{MIT Franka}
        & \textbf{UTD SO101}
        & \textbf{Average} \\
        \midrule

        GVL \cite{ma2024generative}
        & 0.250 & $-0.008$ & 0.292 & 0.056
        & 0.306 & 0.300 & 0.199 \\

        VLAC-2B \cite{zhanggeneralist}
        & 0.292 & 0.167 & $-0.111$ & 0.167
        & $-0.017$ & $-0.033$ & 0.077 \\

        VLAC-8B \cite{zhanggeneralist}
        & 0.271 & 0.064 & $-0.417$ & 0.139
        & 0.072 & 0.167 & 0.049 \\

        RoboDopamine \cite{robodopamine2025general}
        & 0.167 & 0.175 & 0.000 & 0.014
        & 0.220 & 0.067 & 0.107 \\

        Dopamine-GRM-2.0-8B-Preview \cite{robodopamine2025general}
        & 0.479 & 0.442 &  0.333 & 0.431
        & 0.431 & 0.700 & 0.453 \\

        RoboReward-4B \cite{roboreward2026general}
        & 0.625 & 0.332 & 0.333
        & 0.528
        & 0.494
        & 0.700 & 0.502 \\

        RoboReward-8B \cite{roboreward2026general}
        & 0.625 & 0.264 & 0.389 & 0.347
        & 0.396 & 0.767 & 0.465 \\

       Robometer (RoboReward data) \cite{robometer2026scaling}
        & 0.583
        & 0.533
        & 0.646
        & 0.403
        & 0.479
        & 0.667
        & 0.552 \\

        ReWiND \cite{zhang2025rewind}
        & $-0.125$ & 0.336 & 0.028 & $-0.167$
        & 0.080 & $-0.067$ & 0.014 \\

        Robometer (RBM-1M) \cite{robometer2026scaling}
        & 0.646
        & 0.471
        & 0.653
        & 0.694
        & \textbf{0.601}
        & \textbf{0.867}
        & 0.655 \\

        Robometer (Progress only) \cite{robometer2026scaling}
        & 0.083 & 0.231 & 0.333 & 0.389
        & 0.183 & 0.533 & 0.292 \\

        \midrule
        \rowcolor{rynn!15}
        \textbf{RynnValue-4B}
        & 0.542
        & \textbf{0.645}\textsuperscript{\(\dagger\)}
        & \textbf{1.000}\textsuperscript{\(\dagger\)}
        & 0.500
        & 0.518\textsuperscript{\(\dagger\)}
        & \textbf{0.867}\textsuperscript{\(\dagger\)}
        & 0.679 \\
        \rowcolor{rynn!15}
        \textbf{RynnValue-8B}
        & \textbf{0.750}\textsuperscript{\(\dagger\)}
        & 0.504
        & \textbf{1.000}\textsuperscript{\(\dagger\)}
        & \textbf{0.722}\textsuperscript{\(\dagger\)}
        & 0.450
        & 0.800
        & \textbf{0.704}\textsuperscript{\(\dagger\)} \\
        \bottomrule
    \end{tabular}
    }
\end{table*}

As shown in \Cref{tab:rbm_ood_policy_ranking}, RynnValue-8B achieves the highest average Kendall's $\tau_a$ of 0.704, and even the smaller RynnValue-4B reaches 0.679; both exceed the 0.655 obtained by the full Robometer model trained with both progress and trajectory-preference supervision. In terms of the best overall results, RynnValue-8B ranks first on USC Franka and USC xArm. RynnValue-4B ranks first on USC Koch, ties RynnValue-8B at $1.000$ on USC Trossen, and ties Robometer (RBM-1M) at $0.867$ on UTD SO101. Among methods trained without explicit trajectory-level preference supervision, RynnValue variants attain the best results on all six datasets, improving the strongest preference-free prior average of 0.502 to 0.679 at 4B and 0.704 at 8B. RynnValue also substantially outperforms the progress-only Robometer ablation, which achieves an average score of 0.292.
Notably, performance is already strong at 4B and improves only marginally at 8B, suggesting that the gains stem from the temporal-distance formulation and its accompanying training and architectural designs rather than sheer model scale.

These results support temporal distance as an effective supervision target for scaling robotic value models. Unlike explicitly constructed trajectory preferences, temporal-distance labels can be obtained automatically from heterogeneous trajectories while preserving a consistent cost-to-go interpretation across tasks, execution speeds, and robot embodiments. Together with random temporal sampling, order shuffling, and value-isolation attention, this formulation encourages the model to ground its predictions in task-relevant visual evidence rather than dataset-specific progress scales or temporal shortcuts.
Most importantly, RynnValue reaches this quality through a substantially simpler recipe: its targets are read directly from trajectory timestamps, without constructing preference pairs or normalizing every task onto a $[0,1]$ progress scale. That a preference-free, timestamp-derived target can match and even surpass fully preference-supervised training highlights temporal distance as a more scalable route to general-purpose robotic value learning.

\subsection{Instruction-Trajectory Alignment}
\label{sec:instruction_trajectory_alignment}

\begin{figure*}[t]
    \centering
    \includegraphics[width=\textwidth]{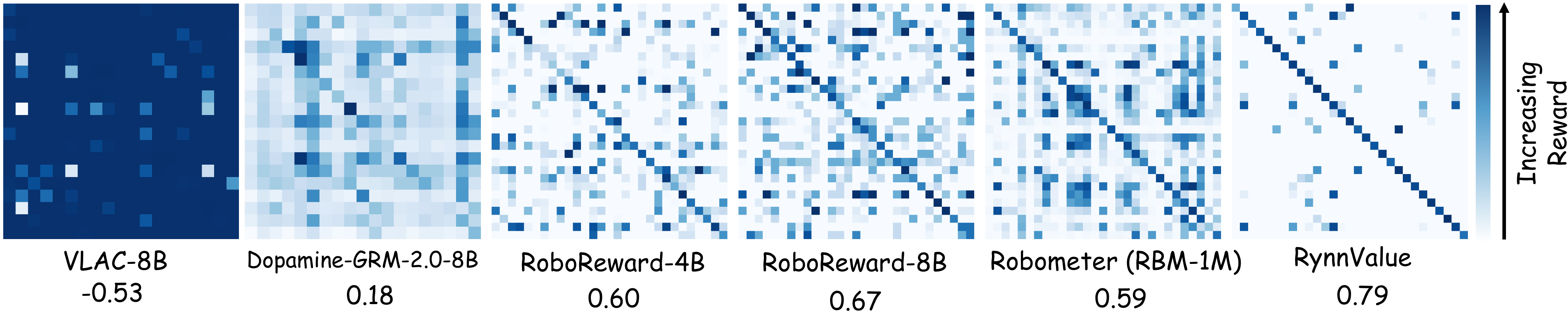}
    \caption{
        Instruction-trajectory confusion matrices.
        Each cell shows the predicted reward when an instruction (rows) is paired with a trajectory (columns); a well-grounded model concentrates mass on the diagonal. Values below each matrix report the normalized diagonal margin. All models are re-evaluated under a unified protocol from their publicly released weights.
    }
    \label{fig:instruction_video_confusion}
\end{figure*}

To assess whether the predicted reward tracks the specified language goal rather than generic visual progress, we perform an instruction-trajectory alignment analysis: we score every instruction against every trajectory and inspect the resulting matrix, following the evaluation protocol of Robometer~\cite{robometer2026scaling}. A well-grounded model should assign higher rewards to matched instruction-trajectory pairs along the diagonal and lower rewards to mismatched pairs.
As shown in \Cref{fig:instruction_video_confusion}, to ensure that all confusion matrices are produced under an identical protocol, we re-evaluate every baseline with its publicly released model weights, and report the resulting diagonal-margin scores.
RynnValue produces the clearest diagonal structure and achieves the highest normalized diagonal margin of \textbf{0.79}, outperforming the strongest baseline at 0.67. In contrast, several baselines exhibit more diffuse off-diagonal responses. These results show that RynnValue more effectively distinguishes matched from mismatched instruction-trajectory pairs, grounding its temporal-distance estimates in the language-specified goal rather than generic visual progress alone.

\subsection{Ablation Study}

\begin{table*}[t]
    \centering
    \caption{
    \textbf{Ablation study on RBM-EVAL-OOD.}
    We report Kendall's $\tau_a$ across six out-of-distribution robot datasets; all variants are evaluated at the same training checkpoint.
    Shuffle denotes temporal-order shuffling, Isolation denotes training-time
    value-isolation attention, Language denotes the auxiliary natural-language supervision,
    Random denotes random temporal sampling, and Relative denotes the relative
    modeling component.
    }
    \label{tab:ablation}
    \setlength{\tabcolsep}{4.5pt}
    \renewcommand{\arraystretch}{1.12}
    \resizebox{\textwidth}{!}{
    \begin{tabular}{lcccccccccccc}
        \toprule
        & \multicolumn{5}{c}{Design Components}
        & \multicolumn{7}{c}{Kendall's $\tau_a$} \\
        \cmidrule(lr){2-6}
        \cmidrule(lr){7-13}
        Variant
        & Shuffle
        & Isolation
        & Language
        & Random
        & Relative
        & \makecell{USC\\Franka}
        & \makecell{USC\\Koch}
        & \makecell{USC\\Trossen}
        & \makecell{USC\\xArm}
        & \makecell{MIT\\Franka}
        & \makecell{UTD\\SO101}
        & Average \\
        \midrule

        w/o Shuffle
        & \xmark & \cmark & \cmark & \cmark & \cmark
        & 0.500 & 0.037 & 0.111 & 0.111 & $-0.104$ & 0.300 & 0.159 \\

        w/o Isolation
        & \cmark & \xmark & \cmark & \cmark & \cmark
        & 0.708 & 0.472 & 0.778 & 0.667 & 0.476 & 0.567 & 0.611 \\

        w/o Language
        & \cmark & \cmark & \xmark & \cmark & \cmark
        & 0.625 & 0.504 & 0.917 & 0.472 & \textbf{0.483} & 0.733 & 0.622 \\

        Uniform Sampling
        & \cmark & \cmark & \cmark & \xmark & \cmark
        & 0.208 & 0.104 & 0.111 & 0.056 & 0.093 & 0.667 & 0.207 \\

        w/o Relative
        & \cmark & \cmark & \cmark & \cmark & \xmark
        & 0.542 & \textbf{0.584} & 0.542 & \textbf{0.778} & 0.420 & 0.600 & 0.578  \\

        \midrule
        \textbf{Full Model (8B)}
        & \cmark & \cmark & \cmark & \cmark & \cmark
        & \textbf{0.750}
        & 0.504
        & \textbf{1.000}
        & 0.722
        & 0.450
        & \textbf{0.800}
        & \textbf{0.704} \\
        \bottomrule
    \end{tabular}
    }
\end{table*}

We ablate the principal training and architectural components of RynnValue in \Cref{tab:ablation}; all variants are evaluated at the same training checkpoint. 
Further ablations in \Cref{app:attention_ablation} compare quantile, fixed-bin, and normalized value tokenizers, as well as inference-time attention mask.
The full model achieves the highest average Kendall's $\tau_a$ of $0.704$. Removing temporal-order shuffling causes the largest degradation, reducing the average score to $0.159$---even turning negative on MIT Franka---as sequence position again becomes a strong proxy for task progress. Replacing random temporal sampling with uniform sampling reduces the average score to $0.207$. The regular intervals it produces let the model exploit a stereotypical value pattern rather than reason from the observed visual content.
Beyond sampling, replacing value-isolation attention with standard eager attention during training decreases performance to $0.611$, confirming that isolation prevents prediction queries from extrapolating values across query groups. Removing the auxiliary natural-language supervision reduces the average score to $0.622$, demonstrating that video-description, instruction-matching, and success supervision provide useful semantic grounding. Finally, removing relative temporal-distance supervision yields a score of $0.578$: by modeling local forward and backward temporal changes, this objective complements the globally anchored absolute prediction and improves the shared visual representation.

\subsection{Scaling Analysis}
\label{sec:scaling_analysis}

\begin{wrapfigure}{r}{0.5\columnwidth}
    \centering
    \vspace{-0.8\baselineskip}
    \includegraphics[width=\linewidth]{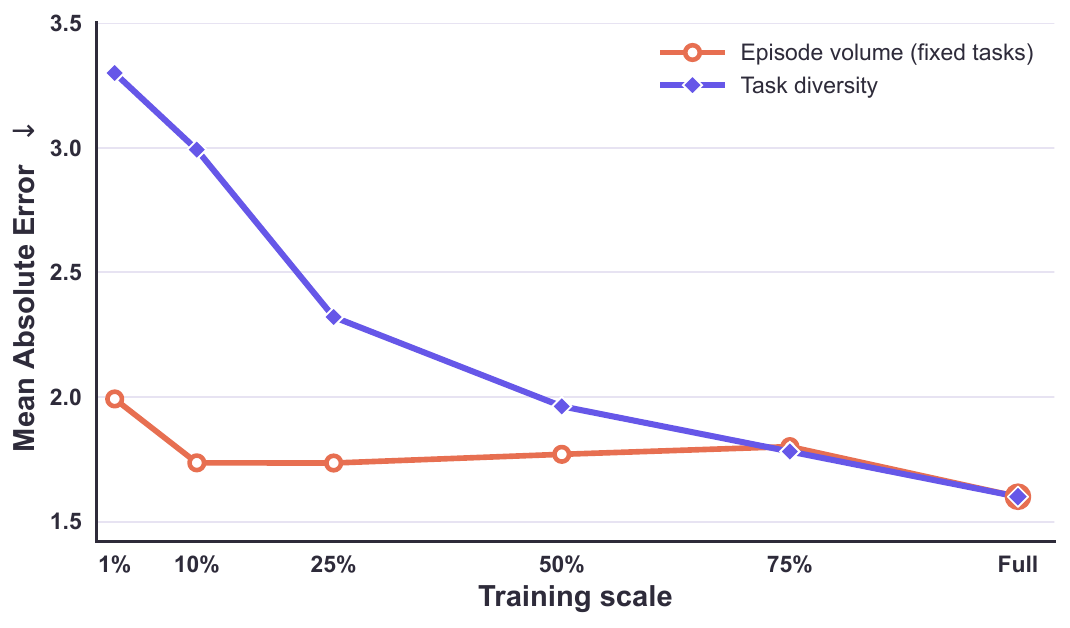}
    \caption{
    Scaling episode volume vs.\ task diversity.
    Mean absolute temporal-distance error on a held-out validation set of unseen tasks.
    We independently scale two aspects of the training set:
    (orange)~episode count with the full task set fixed, and
    (blue)~task count with per-task episode counts fixed.
    Both curves converge to the same full-scale training set at 100\%.
    Task diversity yields a consistently steeper error reduction than episode volume, which saturates early.
    }
    \label{fig:scaling_analysis}
    \vspace{-0.6\baselineskip}
\end{wrapfigure}

Having isolated the contribution of each model component, we next ask whether the heterogeneous data recipe itself drives generalization. To disentangle \emph{quantity} from \emph{diversity}, we construct two families of training subsets from the full mixture. In the \textbf{episode-volume} family (orange), we fix the full set of training tasks and randomly subsample $\{1\%, 10\%, 25\%, 50\%, 75\%\}$ of episodes within each task. In the \textbf{task-diversity} family (blue), we fix per-task episode counts at the full-scale level and randomly subsample the same fractions of tasks. At each fraction, both protocols train on a comparable total number of episodes, ensuring that performance differences reflect diversity rather than data volume. Both families converge to the same full training set at 100\%. We train each variant from scratch under identical optimization settings and report the mean absolute temporal-distance error on a held-out validation set whose tasks do not overlap with any training subset.

As shown in \Cref{fig:scaling_analysis}, episode-volume scaling saturates almost immediately: beyond a small fraction of episodes the error plateaus and does not improve further as more within-task data are added. Task-diversity scaling behaves qualitatively differently: increasing the number of training tasks reduces error monotonically across the entire range, with substantial gains still observable well past the midpoint. This contrast demonstrates that the heterogeneous data recipe contributes not merely additional samples, but the diversity required for general-purpose temporal-value learning: broader task coverage introduces diverse goal structures, visual configurations, and execution patterns that transfer to unseen tasks, whereas repeated observation of the same tasks quickly stops adding signal.

\subsection{Temporal-Value Curve Case Study}
\label{sec:case_study}

\begin{figure*}[t]
    \centering
    \includegraphics[width=\textwidth]{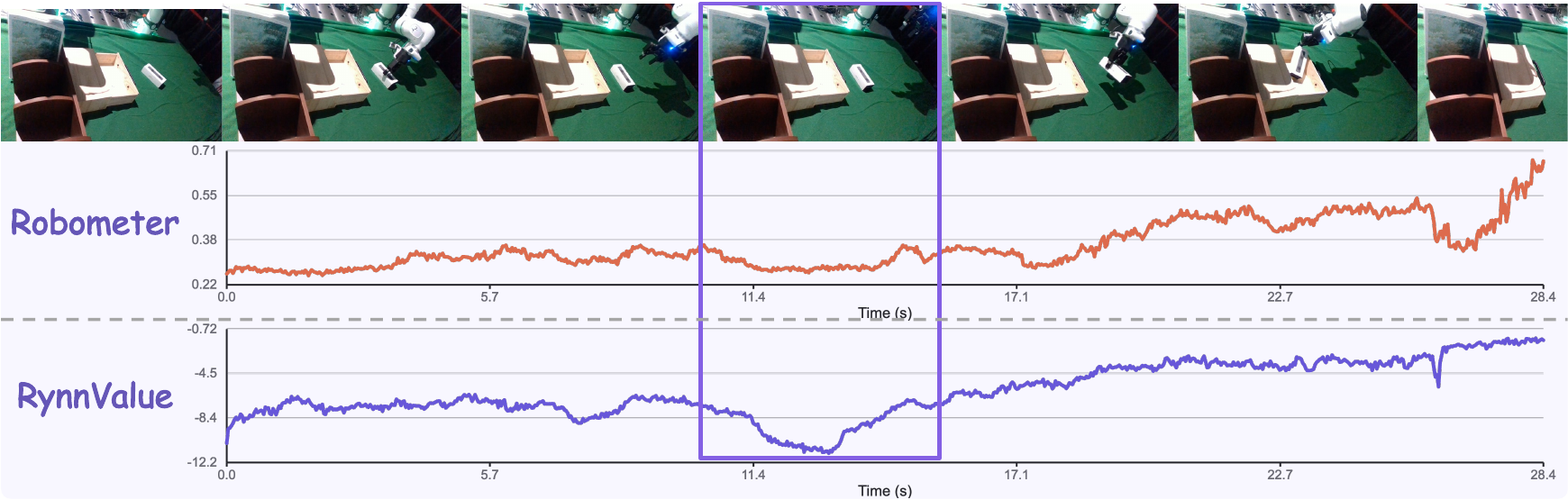}
    \caption{Temporal-value curve comparison on a real-world trajectory. Higher values indicate closer proximity to task completion. The highlighted interval marks a period of task regression where the robot moves away from a productive state; RynnValue responds with a sharp potential drop, whereas Robometer remains relatively flat.}
    \label{fig:qualitative_value_curve}
\end{figure*}

\Cref{fig:qualitative_value_curve} compares the value curves predicted by RynnValue and Robometer along the same real-world trajectory. Because Robometer outputs normalized progress while RynnValue outputs temporal-distance potential ($\Phi_t = -v_t$), the two scales are not directly comparable; we therefore orient both so that higher values indicate closer proximity to task completion. The visualization reveals three notable differences.

\textbf{First}, RynnValue responds more strongly to trial-and-error and task regression. During the highlighted interval, its potential drops substantially as the robot moves away from a productive state, whereas Robometer responds weakly. This sensitivity follows from temporal-order shuffling and value-isolation attention, which ground predictions in visual evidence rather than sequence position or previously exposed values.

\textbf{Second}, after the robot recovers, RynnValue exhibits a steadier increase toward completion, while Robometer contains longer plateaus and abrupt jumps. This behavior benefits from jointly learning absolute and relative temporal values, which capture global remaining cost and local temporal changes, respectively.

\textbf{Third}, RynnValue remains sensitive near task completion: late disturbances cause an immediate potential decrease followed by recovery. This avoids premature reward saturation and better distinguishes intermediate progress from actual completion.

\section{Real-world Policy Learning}
\label{sec:real_world_policy_learning}

Beyond benchmarks, we evaluate whether RynnValue can improve policy learning on a diverse set of challenging real-world manipulation tasks.
Since none of these tasks, their manipulated objects, or the workspace scenes appear in the reward-model training corpus, and no target-domain fine-tuning is performed, RynnValue serves as a zero-shot reward annotator. 
This setting assesses its open-world generalization and its effectiveness for value estimation under physical execution noise.
In \Cref{sec:policy_learning_algorithms}, we additionally provide implementation details on the policy backbones and policy-learning algorithms.

\subsection{Experimental Settings}
\label{sec:experiment_details}

\paragraph{Robot platform.}
Our experiments are conducted on a dual-arm Franka robot system with wrist-mounted cameras on both arms and two additional third-person cameras capturing the workspace from the left and right sides.

\begin{figure*}[t]
    \centering
    \includegraphics[width=\textwidth]{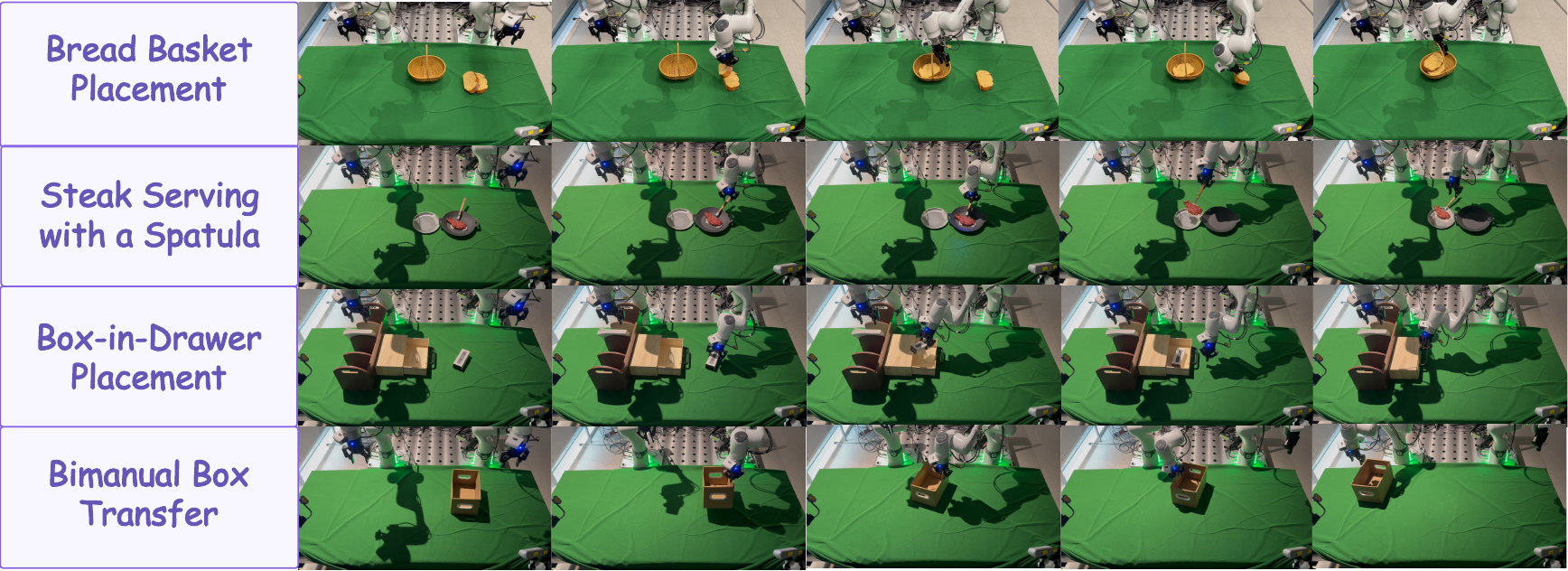}
    \caption{
        \textbf{Representative demonstrations for real-world evaluation.}
        Each row shows a sequence of observations from one manipulation task.
        These tasks cover object grasping, spatial manipulation, and articulated-object interaction.
    }
    \label{fig:real_robot_demonstrations}
\end{figure*}

\paragraph{Tasks and datasets.}
\label{app:real_world_tasks}
We evaluate on the following four real-world robotic manipulation tasks as shown in~\Cref{fig:real_robot_demonstrations}:

\begin{itemize}[labelsep=0.6em, leftmargin=1.2em,itemindent=0em]
    \item \textbf{Bread Basket Placement}: The robot places two pieces of bread into a basket.
    \item \textbf{Steak Serving with a Spatula}: The robot uses a spatula to transfer a steak from a pan to a plate.
    \item \textbf{Box-in-Drawer Placement}: The robot places a box into a drawer and then closes the drawer.
    \item \textbf{Bimanual Box Transfer}: Two robot arms collaboratively move a box to a target location.
\end{itemize}

For each task, we collect approximately 100 successful trajectories and retain every unsuccessful attempt encountered during the same collection process, yielding a mixed-expertise dataset; \Cref{tab:real_world_dataset_statistics} in the appendix reports the task instructions and dataset statistics.

\paragraph{Baselines and reward shaping.}
We compare RynnValue against Robometer, the strongest baseline in our benchmark evaluations, and Sparse Reward, which assigns $-1$ before task completion and $0$ upon successful completion. We additionally include two scalable post-training baselines, multi-task SFT and Filter BC, detailed in \Cref{sec:scalable_post_training}. Plain SFT serves as the per-task imitation-learning reference. To ensure a fair comparison, all reward models are used through the same potential-based reward shaping interface:
\begin{equation}
r'_t
=
\kappa\bigl(\gamma\Phi_{t+1}-\Phi_t\bigr)
+
\begin{cases}
0,  & \text{if $t=T$ and the trajectory is successful}, \\
-1, & \text{otherwise}.
\end{cases}
\end{equation}
where $\Phi_t$ denotes the potential predicted by the reward model at step $t$, $\gamma$ is the discount factor, and $\kappa$ controls the strength of the shaping term. We use $\kappa=0.1$ for RynnValue and $\kappa=1.0$ for Robometer in both the offline and online experiments. We retain the sparse completion term because reward-model predictions may be noisy, whereas the success label provides a clean and reliable signal for the final task objective. These success labels are manually annotated by human operators during both data collection and online policy training. For Sparse Reward, we set $\Phi_t=0$ for all $t$, so the reward is $-1$ before completion and $0$ on the transition that successfully completes the task.

\paragraph{Evaluation protocols.}
\label{app:evaluation_protocols}
We evaluate each policy over 20 trials per tested task and report both the success rate and the average number of action chunks over successful episodes. All offline evaluations use fixed initial configurations. During online evaluation, we randomize the initial object configurations for \textit{Bread Basket Placement}, \textit{Steak Serving with a Spatula}, and \textit{Bimanual Box Transfer} to assess robustness under varying initial conditions, while \textit{Box-in-Drawer Placement} uses a fixed reset because the task requires precise gripper control and box-drawer alignment.

\paragraph{Reset, success, and termination.}
Environment resets are performed by an operator, who repositions the robot and objects and confirms that the next episode may begin. All offline-policy evaluations use fixed initial configurations. During online evaluation, Bread Basket Placement, Steak Serving with a Spatula, and Bimanual Box Transfer use randomized initial-object configurations, whereas Box-in-Drawer Placement uses a fixed reset because the task requires precise gripper control and box--drawer alignment. The operator records a binary task-success label. An episode terminates when success or failure is confirmed or when it reaches the maximum horizon of $600$ low-level control steps. Reward models provide only the shaping potential; they do not determine success or termination.

\paragraph{Offline RL with mixed-expertise datasets.}
First, we evaluate whether RynnValue can recover high-performing policies from mixed-expertise datasets. Specifically, using $\pi_{0.5}$ as the base policy, we apply Implicit Q-Learning (IQL) \cite{kostrikov2021offline} to train policies on these datasets with reward signals from either RynnValue or the baselines.

\paragraph{Online RL with diffusion steering.}
Next, we investigate whether RynnValue can accelerate online policy learning by providing real-time reward specification. We adopt Diffusion Steering via Reinforcement Learning (DSRL) \cite{wagenmaker2025steering} with task-specific policy initialization. Bread Basket Placement and Steak Serving with a Spatula start from their SFT checkpoints, whereas Box-in-Drawer Placement and Bimanual Box Transfer start from the corresponding Robometer-based offline-RL checkpoints. This initialization is held fixed across the online reward variants, whose collected trajectories are specified by either RynnValue or the baselines.

\paragraph{Scalable post-training via data filtering.}
Scalable post-training should improve a generalist policy across \emph{all} tasks at once, rather than requiring a separate adaptation run for each task. We therefore test whether RynnValue can curate multi-task behavior-cloning (BC) data without target-task fine-tuning. Concretely, we pool the demonstrations from all four tasks, score every trajectory segment with RynnValue, and discard the low-quality ones before fine-tuning the pretrained policy.

\paragraph{Inference-time value guidance.}
We evaluate inference-time value guidance on Bread Basket Placement and Steak Serving with a Spatula using a frozen multi-task policy. At each decision step, Cosmos3-Edge samples four candidate action chunks together with their imagined futures, and RynnValue selects the candidate with the largest predicted progress. The sampling and scoring procedure is detailed in \Cref{app:value_guidance}.

\subsection{Reinforcement Learning Results}

\begin{table*}[t]
    \centering
    \small
    \setlength{\tabcolsep}{3.5pt}
    \renewcommand{\arraystretch}{1.15}
    \resizebox{\textwidth}{!}{
    \begin{tabular}{llccccccccc}
        \toprule
        \multirow{2}{*}{\textbf{Algorithm}}
        & \multirow{2}{*}{\textbf{Baseline}}
        & \multicolumn{2}{c}{\textbf{Bread Basket Placement}}
        & \multicolumn{2}{c}{\textbf{Steak Serving with a Spatula}}
        & \multicolumn{2}{c}{\textbf{Box-in-Drawer Placement}}
        & \multicolumn{2}{c}{\textbf{Bimanual Box Transfer}}
        & \multirow{2}{*}{\makecell{\textbf{Average}\\\textbf{Success}}} \\
        \cmidrule(lr){3-4}
        \cmidrule(lr){5-6}
        \cmidrule(lr){7-8}
        \cmidrule(lr){9-10}
        &
        & \makecell{\textbf{Success}\\$\uparrow$}
        & \makecell{\textbf{Avg. Steps}\\$\downarrow$}
        & \makecell{\textbf{Success}\\$\uparrow$}
        & \makecell{\textbf{Avg. Steps}\\$\downarrow$}
        & \makecell{\textbf{Success}\\$\uparrow$}
        & \makecell{\textbf{Avg. Steps}\\$\downarrow$}
        & \makecell{\textbf{Success}\\$\uparrow$}
        & \makecell{\textbf{Avg. Steps}\\$\downarrow$}
        & \\
        \midrule

        \rowcolor{rynn!15}
        \cellcolor{white}\multirow{3}{*}{\makecell{Online RL}}
        & \textbf{RynnValue}
        & \textbf{45.0\%} & $25.9 \pm 8.2$
        & \textbf{75.0\%} & $18.6 \pm 13.1$
        & \textbf{70.0\%} & $\mathbf{27.0 \pm 8.1}$
        & \textbf{100.0\%} & $\mathbf{22.8 \pm 4.7}$
        & \textbf{72.5\%} \\

        & Robometer
        & 35.0\% & $\mathbf{22.7 \pm 5.5}$
        & 45.0\% & $\mathbf{15.2 \pm 2.7}$
        & 65.0\% & $27.7 \pm 5.8$
        & 65.0\% & $25.6 \pm 7.4$
        & 52.5\% \\

        & Sparse
        & 40.0\% & $56.0 \pm 31.7$
        & 45.0\% & $18.4 \pm 4.9$
        & 40.0\% & $27.4 \pm 6.6$
        & 70.0\% & $23.5 \pm 2.7$
        & 48.8\% \\

        \midrule

        \rowcolor{rynn!15}
        \cellcolor{white}\multirow{3}{*}{Offline RL}
        & \textbf{RynnValue}
        & \textbf{100.0\%} & $\mathbf{16.8 \pm 3.1}$
        & \textbf{90.0\%} & $\mathbf{14.9 \pm 4.0}$
        & \textbf{90.0\%} & $\mathbf{14.9 \pm 4.0}$
        & \textbf{50.0\%} & $33.6 \pm 10.5$
        & \textbf{82.5\%} \\

        & Robometer
        & 80.0\% & $18.9 \pm 2.7$
        & 80.0\% & $19.4 \pm 7.2$
        & 50.0\% & $27.3 \pm 6.3$
        & 45.0\% & $\mathbf{28.7 \pm 9.3}$
        & 63.8\% \\

        & Sparse
        & 70.0\% & $26.1 \pm 9.1$
        & 20.0\% & $30.2 \pm 3.3$
        & 0.0\% & --
        & 0.0\% & --
        & 22.5\% \\

        \midrule

        SFT & --
        & 70.0\% & $24.8 \pm 8.0$
        & 25.0\% & $18.6 \pm 6.2$
        & 0.0\% & --
        & 0.0\% & --
        & 23.8\% \\

        \bottomrule
    \end{tabular}
    }
    \caption{
        \textbf{Real-world reinforcement-learning results.}
        We report success rates and the average number of action chunks over
        successful episodes. Average denotes the unweighted mean success rate across four tasks.
    }
    \label{tab:real_robot_rl}
\end{table*}

As shown in \Cref{tab:real_robot_rl}, RynnValue consistently achieves the highest success rate across all four real-world tasks in both online and offline RL. In online OOD evaluation, it reaches an average success rate of $72.5\%$, substantially outperforming Robometer at $52.5\%$ and sparse rewards at $48.8\%$. The improvement is particularly pronounced on \textit{Steak Serving with a Spatula} and \textit{Bimanual Box Transfer}, where RynnValue improves upon Robometer by $30$ and $35$ percentage points, respectively. Under offline RL, RynnValue achieves an average success rate of $82.5\%$, compared with $63.8\%$ for Robometer and only $23.8\%$ for plain SFT.

RynnValue also provides a strong balance between task success and execution efficiency. On \textit{Bread Basket Placement}, it achieves $100\%$ offline success using $16.8$ action chunks on average, compared with $80\%$ and $18.9$ chunks for Robometer and $30\%$ and $30.2$ chunks for multi-task SFT. Similar improvements are observed on \textit{Steak Serving with a Spatula} and \textit{Box-in-Drawer Placement}, where RynnValue reaches $90\%$ success while requiring only $14.9$ chunks. Notably, reinforcement learning with RynnValue successfully solves \textit{Bimanual Box Transfer}, for which the multi-task SFT policy records no successful executions, and lifts \textit{Box-in-Drawer Placement} from $40\%$ to $90\%$ offline. These results show that RynnValue enables substantial improvements over the imitation-learning baseline while maintaining efficient task completion.

The online gains are limited for both reward models on \textit{Box-in-Drawer Placement}. Starting from the shared Robometer-trained offline-RL checkpoint with a $50\%$ success rate, online RL reaches $65\%$ with Robometer and $70\%$ with RynnValue. This task requires precise coordination among the gripper, the box, and the drawer geometry. Because the reward models observe only third-person RGB images, visually similar configurations may correspond to substantially different grasp stability and placement alignment, making it difficult to assign well-calibrated intermediate rewards. Consequently, neither reward model yields a substantial online improvement in this visually ambiguous, precision-sensitive setting, although RynnValue remains slightly more effective than Robometer.

\subsection{Scalable Post-training}
\label{sec:scalable_post_training}

\begin{figure}[t]
    \centering
    \includegraphics[width=\linewidth]{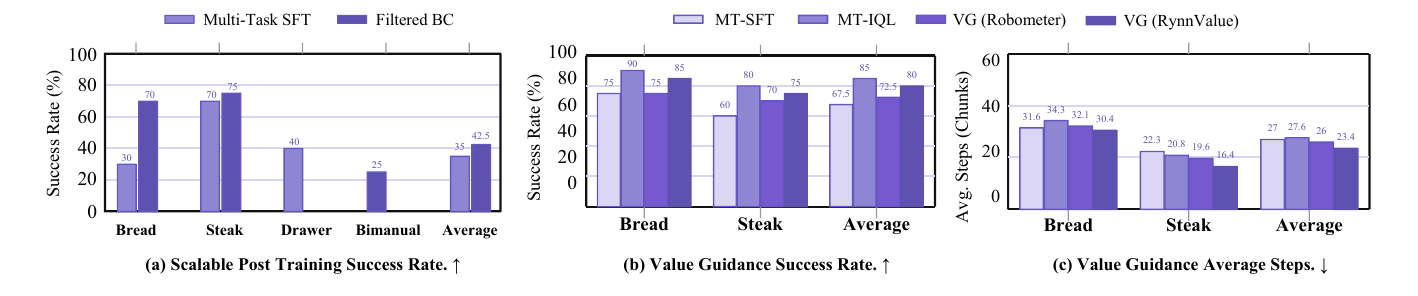}
    \captionsetup{width=\linewidth,skip=\baselineskip}
    \caption{
        \textbf{Multi-task data filtering and value guidance results.}
        (a) Data filtering success rates; missing bars denote tasks with zero successful episodes.
        (b--c) Value guidance success rates and the average number of action chunks over successful episodes, with multi-task BC and multi-task IQL as references.
    }
    \label{fig:scalable_post_training}
    \label{fig:real_robot_value_guided}
\end{figure}

Offline and online RL produce our strongest policies, but both require task-specific training. We therefore consider a task-amortized alternative that trains a single policy jointly across all four tasks. \Cref{fig:scalable_post_training}(a) compares multi-task SFT with Filtered BC. Both variants use the same policy initialization, source dataset, and optimization recipe; the only difference is that Filtered BC removes trajectory segments identified as low quality by RynnValue (please refer to \Cref{app:data_filtering} for implementation details). This controlled design isolates the effect of value-based data selection.
Filtering raises the average success rate from $35.0\%$ to $42.5\%$. The largest gains occur on Bread ($30\%\rightarrow70\%$) and Bimanual ($0\%\rightarrow25\%$), while Steak improves modestly from $70\%$ to $75\%$. The improvement is not uniform: success on Drawer drops from $40\%$ to $0\%$. This failure likely reflects the task's visual ambiguity, as third-person RGB observations alone may not capture the grasp stability or box--drawer alignment required for precise manipulation, causing useful segments to be incorrectly filtered. Overall, filtering the pooled data with RynnValue, without any target-task adaptation, produces the strongest BC policy after a single multi-task training run. This result shows that RynnValue can support both reward shaping and scalable data curation, making it an effective and practical interface for post-training generalist robot policies.

\subsection{Inference-time Value Guidance}
\label{sec:value_guided_inference}

Inference-time value guidance---which leverages a well-trained state--action value function to steer the action proposals of a \emph{frozen} policy---is another important application of RynnValue. We instantiate this idea on Cosmos3-Edge-2B~\cite{nvidia2026cosmos3} (\Cref{app:cosmos_edge_policy}), a performant world-action model. Specifically, we take the multi-task BC policy (\textsc{MT-SFT}) as the frozen proposal generator and rescore its action candidates with Robometer and RynnValue, yielding \textsc{VG (Robometer)} and \textsc{VG (RynnValue)}, respectively. For reference, we additionally train a multi-task variant of IQL (\textsc{MT-IQL}), which we regard as an upper bound on what these training-free approaches can achieve. \Cref{fig:real_robot_value_guided}(b--c) supports three observations. First, when the scorer is the only difference, replacing Robometer with RynnValue raises average success from $72.5\%$ to $80.0\%$ while cutting the mean length of successful episodes from $26.0$ to $23.4$ action chunks. Second, and more tellingly, the gain over the unguided \textsc{MT-SFT} policy amounts to $12.5$ percentage points for RynnValue but only $5.0$ points for Robometer, indicating that the benefit of best-of-$N$ selection hinges on the fidelity of the scorer rather than on the candidate budget itself. Finally, \textsc{VG (RynnValue)} comes within $5$ points of \textsc{MT-IQL} on both tasks, all without updating either the policy or the value model. Taken together, these results show that RynnValue offers a markedly more effective value-ranking signal for inference-time action selection.

\section{Conclusion and Future Works}
\noindent

We presented RynnValue, a value foundation model for robotic manipulation that replaces trajectory-internal progress with temporal distance as its supervision target. By treating a state's directed, goal-conditioned cost-to-go as the learning objective, RynnValue turns heterogeneous data into a single, preference-free value interface: labels are derived directly from timestamps, and disparate embodiments, viewpoints, and task durations are unified without dataset-specific progress normalization. To make temporal-value learning robust at scale, we combined random temporal sampling and temporal-order shuffling with value-isolation attention, suppressing the sampling-interval, sequence-order, and value-extrapolation shortcuts that otherwise leave learned values insensitive to failures and regressions.

Trained without a single preference label, RynnValue surpasses the fully preference-supervised state of the art on RBM-EVAL-OOD and more than doubles a progress-only counterpart, while generalizing zero-shot across unseen tasks, embodiments, and viewpoints. As a reusable value interface, it improves both online and offline real-world policy learning through potential-based shaping, curates multi-task demonstration data through value-based filtering, and steers frozen policies through inference-time candidate selection. Together, these results indicate that temporal distance is both a scalable supervision target for value foundation models and a practical reward interface for generalist robot policies.

Looking forward, we aim to extend RynnValue toward broader temporal horizons, richer value semantics, and more diverse embodiments. RynnValue currently estimates temporal distance from a short window of sampled observations, so extending it to longer horizons and streaming inference would broaden its use as an online reward source. Its targets assume an approximately minimum-time objective, and incorporating task-specific costs such as energy, safety, or precision could yield richer value semantics. Finally, we plan to scale RynnValue to a wider range of end-effectors, including dexterous hands, as well as mobile manipulation settings, where long-horizon navigation and interaction must be jointly grounded in the same temporal-value interface.

\bibliographystyle{assets/plainnat}
\bibliography{paper.bib}

\newpage
\beginappendix

\phantomsection
\addcontentsline{toc}{section}{A. Appendix} 

\addtocontents{toc}{\protect\setcounter{tocdepth}{0}}

\appendix
\renewcommand{\thesection}{\Alph{section}}
\renewcommand{\thesubsection}{\Alph{subsection}}

\subsection{RynnValue Dataset}
\label{sec:dataset_details}
This section complements~\Cref{sec:rynnvalue_dataset} with the source-aware curation pipeline applied before constructing temporal-distance targets. Raw robot corpora do not always pair each trajectory segment with a meaningful, task-relevant language goal: their annotations may contain placeholders, data-quality tags, truncated commands, or descriptions of pure robot motion. We therefore curate Open X-Embodiment (OXE), InternData-A1 (InternA1), Galaxea Open-World (Galaxea), and RoboCOIN in a source-specific manner. Since OXE and InternA1 are counted in episodes, whereas Galaxea and RoboCOIN are counted in annotated subtask segments, we refer to their entries collectively as \emph{trajectory units}.

\paragraph{Instruction validation.}
We first retain only instructions that specify a well-formed and actionable robot goal. Source-specific usable-instruction sets are used when canonical task vocabularies are available, while malformed strings, placeholders, data-quality metadata, and ambiguous noun-only labels are removed. We also discard CJK annotations from sources whose task interface is defined in English. This stage removes annotation artifacts without altering the semantics of valid manipulation tasks.

\paragraph{Action-relevance filtering.}
Some mobile-manipulation trajectories contain segments that describe only locomotion or approach behavior, without a manipulation objective. This issue is most prominent in Galaxea. We identify such segments when motion predicates such as moving, approaching, turning, or walking occur without any manipulation predicate, and remove them from temporal-distance training. This ensures that a retained segment describes observable progress toward an interaction goal rather than incidental platform motion.

\Cref{tab:data_curation_examples} gives representative examples of the annotation issues addressed by the pipeline, spanning several qualitatively different failure modes: malformed language, dataset metadata, task-irrelevant motion.

\begin{table*}[h]
    \centering
    \caption{
        \textbf{Representative annotation issues addressed by data curation.}
        Non-English examples are described rather than reproduced verbatim to
        avoid introducing non-Roman fonts into the manuscript.
    }
    \label{tab:data_curation_examples}
    \small
    \setlength{\tabcolsep}{6pt}
    \renewcommand{\arraystretch}{1.12}
    \resizebox{\textwidth}{!}{
    \begin{tabular}{l l l}
        \toprule
        \textbf{Issue}
        & \textbf{Representative annotation}
        & \textbf{Source} \\
        \midrule
        Non-English task annotation
        & Chinese-language manipulation instruction
        & Galaxea \\
        Placeholder or truncated label
        & \texttt{P}, \texttt{shirts}, \texttt{undefined}
        & InternA1 \\
        Data-quality metadata
        & \texttt{no robot motion}, \texttt{skip frame}
        & OXE \\
        Pure-motion instruction
        & \texttt{move to the table}, \texttt{approach the cabinet}
        & Galaxea \\
        \bottomrule
    \end{tabular}
    }
\end{table*}

\Cref{tab:data_curation_stats} summarizes the quantitative effect of the complete pipeline. Overall, curation retains 1,436,150 of 1,722,966 trajectory units (83.35\%) while preserving 192,989 of 194,967 unique instructions (98.99\%). This gap between unit retention and instruction retention indicates that the removed volume is concentrated in invalid annotations rather than in rare valid tasks.

\begin{table*}[h]
    \centering
    \caption{
        \textbf{Summary of source-specific data curation.}
        OXE and InternA1 are counted in episodes, while Galaxea and RoboCOIN
        are counted in annotated subtask segments. The total therefore denotes
        an aggregate number of trajectory units rather than a homogeneous
        episode count.
    }
    \label{tab:data_curation_stats}
    \small
    \setlength{\tabcolsep}{4.5pt}
    \renewcommand{\arraystretch}{1.12}
    \resizebox{\textwidth}{!}{
    \begin{tabular}{l r r r r r r}
        \toprule
        \textbf{Source}
        & \textbf{Raw units}
        & \textbf{Curated units}
        & \textbf{Unit retention}
        & \textbf{Raw instructions}
        & \textbf{Curated instructions}
        & \textbf{Instruction retention} \\
        \midrule
        OXE (episode)
        & 961,253 & 693,037 & 72.10\%
        & 180,290 & 180,090 & 99.89\% \\
        InternA1 (episode)
        & 320,910 & 320,905 & 99.99\%
        & 350 & 348 & 99.43\% \\
        Galaxea (segment)
        & 97,287 & 78,692 & 80.89\%
        & 12,685 & 10,909 & 86.00\% \\
        RoboCOIN (segment)
        & 343,516 & 343,516 & 100.00\%
        & 1,642 & 1,642 & 100.00\% \\
        \midrule
        \textbf{Total}
        & \textbf{1,722,966} & \textbf{1,436,150}
        & \textbf{83.35\%}
        & \textbf{194,967} & \textbf{192,989}
        & \textbf{98.99\%} \\
        \bottomrule
    \end{tabular}
    }
\end{table*}

The resulting mixture retains broad task coverage while reducing several sources of supervision bias. Under a diagnostic first-verb grouping, the miscellaneous \emph{other} category decreases from 4.54\% to 1.37\%, reflecting the removal of malformed and non-actionable annotations. The dominant \emph{pick-place} category decreases from 43.38\% to 33.61\% after filtering invalid OXE annotations. The three largest skill groups still account for 68.21\% of the curated data, showing that the pipeline does not impose an artificial uniform distribution; instead, it improves task relevance and limits template domination while preserving the natural structure of the robot data.

As an additional post-hoc diagnostic, we normalize the first verb of each instruction into an atomic skill, merging morphological variants such as \emph{pickup}, \emph{picking}, and \emph{picked}. The number of atomic verb categories decreases from 640 in the raw annotations to 549 after curation. The removed categories correspond mainly to empty tokens, termination markers, malformed strings, and other annotation artifacts. This atomic-skill analysis is used only to diagnose the effect of data curation; it does not define the temporal-distance targets or alter the training sampler.

\subsection{Model Details}
\label{sec:model_details}
This section complements~\Cref{sec:model_architecture} with the construction of the frozen temporal-distance quantile supports used by RynnValue's distributional temporal heads, together with their empirical statistics on the training mixture.

\subsubsection{Temporal-Distance Quantile Support Construction}
\label{app:temporal_quantile_supports}

We construct the temporal heads' quantile supports once from training-data metadata and keep them fixed throughout training and inference. For frame index $t$ in a trajectory with relabeled completion cutoff $c$ and frame rate $f$, we include the following non-negative label in the support-fitting pool:
\begin{equation}
    y_t = \max\!\left(\frac{c-t}{f}, 0\right),
    \label{eq:absolute_temporal_label}
\end{equation}
This label measures the remaining time to the relabeled completion cutoff in seconds. We compute it for every training frame, yielding a frame-weighted empirical distribution in which longer trajectories contribute proportionally more samples. We use all available training trajectories without imposing a maximum-duration threshold.

Let $\mathcal{Y}$ be the resulting multiset of $N=2{,}167{,}034{,}629$ frame-level labels. For $B=256$ support centers, we define the $i$-th non-negative knot as
\begin{equation}
    b_i = \widehat{Q}_{\mathcal{Y}}\!\left(\frac{i}{B-1}\right),
    \qquad i\in\{0,\ldots,B-1\},
    \label{eq:temporal_quantile_knots}
\end{equation}
where $\widehat{Q}_{\mathcal{Y}}$ is the empirical quantile function. We store the knots in float32. If consecutive quantile levels yield the same float32 value, we advance the later knot to the next representable value to maintain a strictly increasing support. We do not clip the upper tail, so the final knot equals the maximum observed remaining time of $636.33$ seconds. The signed relative support uses the same non-negative temporal scale and the symmetric construction described in \Cref{sec:model_architecture}.

\begin{table}[H]
    \centering
    \caption{\textbf{Temporal-distance support statistics.} The distribution statistics are computed over all frame-level labels used to fit the frozen non-negative support. Spacing statistics are computed between consecutive quantile knots.}
    \label{tab:temporal_quantile_stats}
    \small
    \setlength{\tabcolsep}{4.5pt}
    \renewcommand{\arraystretch}{1.12}
    \begin{tabular}{@{}lr@{}}
        \toprule
        \textbf{Statistic} & \textbf{Value} \\
        \midrule
        Number of frame-level labels & $2{,}167{,}034{,}629$ \\
        Number of quantile knots & $256$ \\
        Minimum / maximum label & $0.00$ / $636.33$ s \\
        Mean label & $25.82$ s \\
        50th / 90th / 99th percentile & $11.57$ / $68.73$ / $181.07$ s \\
        Minimum adjacent-knot spacing & $0.0333$ s \\
        Median adjacent-knot spacing & $0.1667$ s \\
        Maximum adjacent-knot spacing & $419.30$ s \\
        \bottomrule
    \end{tabular}
\end{table}

The empirical quantile transform concentrates support centers at short horizons, where frame-level labels are densest. As quantified in \Cref{tab:temporal_quantile_allocation}, nearly half of the centers lie within the first 10 seconds, and $84.8\%$ lie within 50 seconds. The sparse upper tail has correspondingly coarser resolution: the final adjacent-knot interval spans $217.03$ to $636.33$ seconds and is the widest interval in the support.

\begin{table}[H]
    \centering
    \caption{\textbf{Allocation of quantile knots over temporal horizons.} Counts are cumulative and include all support centers no greater than the stated horizon.}
    \label{tab:temporal_quantile_allocation}
    \small
    \setlength{\tabcolsep}{4.5pt}
    \renewcommand{\arraystretch}{1.12}
    \begin{tabular}{@{}rrr@{}}
        \toprule
        \textbf{Horizon} & \textbf{Knot count} & \textbf{Support share} \\
        \midrule
        $\leq 1$ s   & $19$  & $7.4\%$ \\
        $\leq 5$ s   & $75$  & $29.3\%$ \\
        $\leq 10$ s  & $119$ & $46.5\%$ \\
        $\leq 20$ s  & $165$ & $64.5\%$ \\
        $\leq 50$ s  & $217$ & $84.8\%$ \\
        $\leq 100$ s & $241$ & $94.1\%$ \\
        $\leq 200$ s & $254$ & $99.2\%$ \\
        \bottomrule
    \end{tabular}
\end{table}

Each knot index defines a fixed output-bin identity. We use the same frozen centers to construct two-hot training targets and decode predictions as expected bin centers. Changing the knot locations would therefore alter the meaning of the output dimensions even if the bin count stayed the same. Accordingly, the support remains fixed across training stages, fine-tuning runs, and checkpoint evaluation.

\subsection{Reward Evaluation Details}
\label{sec:eval_details}
This section complements~\Cref{sec:benchmark_eval} with the RBM-EVAL-OOD inference protocol and further ablations on value tokenizers and inference-time attention masks. A paired comparison of attention implementations quantifies the effect of matching the inference mask to training, while the tokenizer comparison examines how the choice of temporal-distance support affects trajectory-ranking quality.

\subsubsection{RBM-EVAL-OOD Evaluation Protocol}
\label{app:rbm_ood_protocol}
\label{app:eval_baselines}
The methods in \Cref{tab:rbm_ood_policy_ranking} cover prompted value estimation, progress prediction, and trajectory-level reward learning. Their reported ranking scores are taken from Robometer~\cite{robometer2026scaling}; the instruction-trajectory analysis in \Cref{fig:instruction_video_confusion} re-evaluates the subset shown there from released weights.

\paragraph{GVL.}
Generative Value Learning (GVL)~\cite{ma2024generative} prompts a pretrained vision-language model to estimate task progress from video frames without reward-model fine-tuning. The benchmark implementation shuffles the sampled frames before prompting and uses GPT-5-mini to produce progress estimates.

\paragraph{VLAC.}
VLAC~\cite{zhanggeneralist} jointly learns action generation and a language-conditioned progress critic from human and robot trajectories. Its critic compares two observations and predicts their relative progress. We report both released model sizes, VLAC-2B and VLAC-8B.

\paragraph{Robo-Dopamine.}
Robo-Dopamine~\cite{robodopamine2025general} learns a process reward from fine-grained progress changes between observations, including forward and rewound frame pairs. The table includes the original RoboDopamine checkpoint and the later Dopamine-GRM-2.0-8B-Preview checkpoint; these are distinct model versions.

\paragraph{RoboReward.}
RoboReward~\cite{roboreward2026general} predicts a discrete task-execution score from a trajectory and instruction. Its supervision is at the trajectory level rather than a dense within-trajectory progress target. We report its 4B and 8B checkpoints separately.

\paragraph{ReWiND.}
ReWiND~\cite{zhang2025rewind} learns language-conditioned progress from visual and text features and uses rewound video segments to expose task regression. The benchmark trains this baseline on RBM-1M for the reported comparison.

\paragraph{Robometer.}
Robometer~\cite{robometer2026scaling} predicts frame-level progress and success while learning to compare which of two trajectories better satisfies an instruction. The ``RoboReward data'' and ``RBM-1M'' rows distinguish its training data; the ``Progress only'' row removes trajectory-comparison supervision and isolates progress-based training. These variants should not be conflated with the separate RoboReward model above.

We use Robometer's processed trajectory inputs and trajectory-ranking evaluator for RBM-EVAL-OOD~\cite{robometer2026scaling}.

\paragraph{Metric and scoring.}
We use the absolute temporal-distance head to obtain the ranking signal. Let $v_{\mathrm{end}}(\tau,\ell)$ be the predicted remaining time at the final queried observation of trajectory $\tau$ under instruction $\ell$. Following the potential definition in \Cref{eq:rv_potential}, its ungated score is $-v_{\mathrm{end}}$, so larger scores indicate that the trajectory ends closer to task completion. We enable instruction-mismatch gating: when the language branch predicts \texttt{Match: No}, the evaluator replaces the trajectory score with $-100$. The resulting score is
\begin{equation}
    s(\tau,\ell)=
    \begin{cases}
        -100, & \text{if the model predicts \texttt{Match: No}}, \\[2pt]
        -v_{\mathrm{end}}(\tau,\ell), & \text{otherwise}.
    \end{cases}
    \label{eq:rbm_ood_score}
\end{equation}
Relative temporal predictions are not fused into this score. Neither the generated video description nor the success judgment contributes an additional scoring term; the matching judgment is used only for the mismatch gate. Within each task, we compute Kendall's $\tau_a$ between predicted scores and the annotated quality ordering, average over tasks within each suite, and report the unweighted mean over the six suites. We additionally report $\tau_{\mathrm{rw}}$, the same ranking statistic obtained with rewind-augmented scoring.

\subsubsection{Ablation Studies}
\label{app:attention_ablation}
\Cref{tab:rbm_ood_attention_tau} compares value tokenizers and inference-time attention on RBM-EVAL-OOD. All models were trained with value-isolation attention; paired rows change only the inference mask on the same checkpoint and 567 trajectories. This differs from the 976-trajectory benchmark in \Cref{tab:rbm_ood_policy_ranking} and the training-time ablation in \Cref{tab:ablation}.

\begin{table*}[h]
    \centering
    \caption{
    \textbf{Ablation of value tokenizers and attention implementations.}
    All models were trained with value-isolation attention; the Attention column specifies the inference-time mask.
    We apply the causal mask with Hugging Face Transformers' built-in scaled dot-product attention (SDPA) backend (\texttt{attn\_implementation="sdpa"}).
    Paired attention runs use the same checkpoint and the same 567 processed trajectories.
    Value tokenizer is the support used for the temporal-distance head.
    Fixed-bin rows replace the quantile-knot support with fixed-interval bins.
    Average is the unweighted mean over suites.
    Bold values indicate the best result in each suite, including ties.
    }
    \label{tab:rbm_ood_attention_tau}
    \setlength{\tabcolsep}{5pt}
    \renewcommand{\arraystretch}{1.15}
    \renewcommand{\cellalign}{cc}
    \resizebox{\textwidth}{!}{
    \begin{tabular}{*{10}{c}}
        \toprule
        \multicolumn{3}{c}{Design Components}
        & \multicolumn{7}{c}{Kendall's $\tau_a$} \\
        \cmidrule(lr){1-3}
        \cmidrule(lr){4-10}
        \makecell{Parameters}
        & \makecell{Value\\Tokenizer}
        & \makecell{Attention}
        & \makecell{USC\\Franka}
        & \makecell{USC\\Koch}
        & \makecell{USC\\Trossen}
        & \makecell{USC\\xArm}
        & \makecell{MIT\\Franka}
        & \makecell{UTD\\SO101}
        & \makecell{Average} \\
        \midrule

        8B
        & quantile & value-isolation
        & 0.667 & 0.437 & 0.972 & 0.556 & 0.442 & 0.800 & 0.646 \\

        4B
        & quantile & value-isolation
        & 0.583 & 0.485 & 0.931 & 0.500 & 0.450 & 0.733 & 0.614 \\

        8B
        & normalized & value-isolation
        & 0.583 & 0.416 & 0.778 & 0.500 & 0.493 & 0.733 & 0.584 \\

        8B
        & normalized & causal mask
        & 0.583 & 0.635 & 0.806 & 0.639 & 0.475 & 0.867 & 0.667 \\

        8B
        & fixed bins & causal mask
        & 0.667 & 0.544 & \textbf{1.000} & 0.500 & 0.503 & 0.833 & 0.675 \\

        4B
        & fixed bins & causal mask
        & 0.542 & 0.488 & 0.917 & 0.667 & 0.473 & \textbf{0.933} & 0.670 \\

        \midrule
        8B
        & quantile & causal mask
        & \textbf{0.750} & 0.504 & \textbf{1.000} & \textbf{0.722} & 0.450 & 0.800 & \textbf{0.704} \\

        4B
        & quantile & causal mask
        & 0.542 & \textbf{0.645} & \textbf{1.000} & 0.500 & \textbf{0.518} & 0.867 & 0.679 \\
        \bottomrule
    \end{tabular}
    }
\end{table*}

\paragraph{Attention.}
Isolation lets value-query tokens interact within a prediction group but blocks their representations from later observation and value groups. The causal mask opens that path at inference, allowing later queries to use earlier value-query states, though not their decoded scalar predictions. On paired trajectories, the causal mask raises mean Kendall's $\tau_a$ from $0.646$ to $0.704$ for 8B quantile, $0.614$ to $0.679$ for 4B quantile, and $0.584$ to $0.667$ for 8B normalized. The effects vary by suite: the 8B quantile gain is $0.166$ on USC xArm but only $0.008$ on MIT Franka; 4B quantile gains $0.160$ on USC Koch yet loses $0.041$ on USC Franka. Earlier value states may disambiguate the final queried score by carrying trajectory context. Terminal-score ranking cannot, however, establish better frame-level distance estimates or sensitivity to local regression. The ``w/o Isolation'' model in \Cref{tab:ablation} instead removes the mask during training and lowers the reported mean from $0.704$ to $0.611$: unlike the inference switch here, that intervention permits the cross-group shortcut throughout learning.

\paragraph{Value tokenizer.}
The strongly skewed training labels (median $11.57$ s, 90th percentile $68.73$ s, maximum $636.33$ s; \Cref{tab:temporal_quantile_stats}) motivate non-uniform temporal resolution. Of $256$ quantile centers, $119$ fall within $10$ s and $217$ within $50$ s, versus about $20$ of the same $256$ fixed-interval centers within $50$ s when spread to the same maximum (\Cref{tab:temporal_quantile_allocation}). Two-hot targets thus resolve short horizons finely but coarsen the sparse tail. Normalized progress changes the target scale: the same episode fraction can represent different remaining durations, weakening a shared cost-to-go interpretation. Consistent with these mechanisms, 8B quantile exceeds normalized under isolation ($0.646$ versus $0.584$) and both normalized and fixed bins with a causal mask ($0.704$ versus $0.667$ and $0.675$); at 4B, its mean with a causal mask is $0.679$ versus $0.670$ for fixed bins. The suite-level reversals qualify this mean advantage: normalized 8B leads on USC Koch; fixed-bin 4B leads on USC xArm and UTD SO101; quantile 4B leads on USC Koch and USC Trossen. Fixed bins may help at long horizons; normalized progress may help when fraction completed matters. The table lacks pairwise horizons and errors needed to test either explanation, so label density remains a plausible account of the mean rather than a demonstrated cause of each suite-level difference.

\subsection{Real-World Task and Dataset Statistics}
\label{app:real_world_task_stats}

This section complements the real-world task definitions in \Cref{sec:experiment_details}. \Cref{tab:real_world_dataset_statistics} reports the language instructions and dataset statistics of the four real-world manipulation tasks; for each task, approximately 100 successful trajectories are collected and every unsuccessful attempt during the same collection process is retained, yielding a mixed-expertise dataset.

\begin{table}[H]
    \centering
    \caption{\textbf{Real-world task instructions and dataset statistics.} Success rate is computed over all collected trajectories for each task.}
    \label{tab:real_world_dataset_statistics}
    \small
    \setlength{\tabcolsep}{4pt}
    \renewcommand{\arraystretch}{1.15}
    \begin{tabularx}{\textwidth}{
        @{}
        >{\raggedright\arraybackslash}p{0.24\textwidth}
        >{\raggedright\arraybackslash}X
        ccc
        >{\centering\arraybackslash}p{0.10\textwidth}
        @{}
    }
        \toprule
        \multirow{2}{*}{\textbf{Task}}
        & \multirow{2}{*}{\textbf{Language instruction}}
        & \multicolumn{3}{c}{\textbf{Trajectories}}
        & \multirow{2}{*}{\makecell{\textbf{Success}\\\textbf{rate}}} \\
        \cmidrule(lr){3-5}
        &
        & \textbf{Success}
        & \textbf{Failure}
        & \textbf{Total}
        & \\
        \midrule
        Bread Basket Placement
        & ``Put the two pieces of bread in the basket.''
        & 99 & 4 & 103 & 96.1\% \\
        \addlinespace[2pt]
        Steak Serving with a Spatula
        & ``Move the steak from the pan to the plate.''
        & 98 & 4 & 102 & 96.1\% \\
        \addlinespace[2pt]
        Box-in-Drawer Placement
        & ``Put the box in the drawer and close it.''
        & 101 & 3 & 104 & 97.1\% \\
        \addlinespace[2pt]
        Bimanual Box Transfer
        & ``Move the box from the right side to the left side.''
        & 100 & 1 & 101 & 99.0\% \\
        \midrule
        \textbf{Total}
        & --
        & \textbf{398}
        & \textbf{12}
        & \textbf{410}
        & \textbf{97.1\%} \\
        \bottomrule
    \end{tabularx}
\end{table}

\subsection{Policy Learning Algorithms}
\label{sec:policy_learning_algorithms}
This section complements~\Cref{sec:real_world_policy_learning} with the policy backbones used on the real-world platform and the offline IQL and online DSRL procedures that consume them.

\subsubsection{Policy Backbones}
\label{app:policy_backbones}

Real-world policy learning uses two flow-matching backbones of different types: a vision-language-action (VLA) model and a world-action model (WAM). Offline IQL, online DSRL, per-task SFT, and multi-task data filtering all fine-tune or decode from the VLA, $\pi_{0.5}$. Inference-time value guidance instead uses a Cosmos3-Edge WAM jointly fine-tuned on all four real-world tasks to generate future video and actions; its value-guidance evaluation is restricted to Bread Basket Placement and Steak Serving with a Spatula. Both backbones are conditioned on multi-view RGB and a language instruction; neither receives proprioception. They differ in how they consume vision and emit actions: $\pi_{0.5}$ ingests per-camera $224\times224$ views, predicts a 16-step chunk padded to width 32, and uses relative gripper commands at $10$\,Hz, whereas Cosmos3-Edge packs cameras into a single mosaic, jointly generates 16 future frames and 16 absolute-joint actions padded to width 64, and is trained at $30$\,FPS.

\paragraph{$\pi_{0.5}$ Backbone.}
\label{app:shared_policy_backbone}
SFT, offline IQL, and online DSRL all use the same flow-matching vision-language-action policy, $\pi_{0.5}$. Sharing this backbone isolates the effect of the learning algorithm and reward specification.
The VLA consumes camera images resized to $224\times224$, image-validity masks, the task instruction, and the flow-matching timestep. It outputs an action chunk with horizon $H=16$ and padded action dimension $d_a=32$.
The low-level controller executes absolute joint-position commands and relative gripper commands at $10$\,Hz. A single-arm action contains seven absolute joint-position values and one relative gripper value; a bimanual action contains 14 absolute joint-position values and 2 relative gripper values. These task-specific actions are padded to $d_a=32$. Each VLA prediction contains an action chunk with horizon $H=16$.

\paragraph{Cosmos3-Edge Backbone.}
\label{app:cosmos_edge_policy}
The inference-time value-guidance experiments in \Cref{sec:value_guided_inference} use a Cosmos3-Edge world-action model (WAM) jointly fine-tuned on all four real-world tasks, rather than the $\pi_{0.5}$ backbone of the IQL and DSRL runs. We initialize from Cosmos3-Edge-Policy-DROID, retain the Nemotron-2B reasoner and a mixture-of-transformers (MoT) generation branch, and jointly synthesize future video and action chunks by flow matching. At inference the policy is queried in world-action mode: given the current multi-view observation and the language instruction, it generates $N=4$ candidate future videos together with the corresponding 16-step action chunks, which RynnValue then ranks. We record the interfaces and training recipe of that policy below.

\paragraph{Tasks and data mixture.}
We jointly fine-tune on the same four real-world tasks as in \Cref{app:real_world_tasks}, sampling them in equal proportion $1{:}1{:}1{:}1$ so that each task occupies approximately 25\% of every training batch. Failed episodes are excluded. \Cref{tab:cosmos_edge_tasks} lists the embodiment, action dimension, and camera streams of each task. Bread Basket Placement, Steak Serving with a Spatula, and Box-in-Drawer Placement are single-arm skills; Bimanual Box Transfer is a dual-arm skill.

\begin{table}[H]
    \centering
    \caption{\textbf{Four-task Cosmos3-Edge SFT mixture.} Tasks are sampled in equal proportion. Failed episodes are not used.}
    \label{tab:cosmos_edge_tasks}
    \small
    \setlength{\tabcolsep}{6pt}
    \renewcommand{\arraystretch}{1.15}
    \begin{tabularx}{\textwidth}{
        @{}
        >{\raggedright\arraybackslash}p{0.28\textwidth}
        >{\raggedright\arraybackslash}p{0.18\textwidth}
        c
        >{\raggedright\arraybackslash}X
        @{}
    }
        \toprule
        \textbf{Task}
        & \textbf{Embodiment}
        & \textbf{Action dim.}
        & \textbf{Visual input} \\
        \midrule
        Bread Basket Placement
        & single-arm
        & 8
        & \texttt{left\_side}, \texttt{left\_wrist} \\
        \addlinespace[2pt]
        Steak Serving with a Spatula
        & single-arm
        & 8
        & \texttt{left\_side}, \texttt{left\_wrist} \\
        \addlinespace[2pt]
        Box-in-Drawer Placement
        & single-arm
        & 8
        & \texttt{left\_side}, \texttt{left\_wrist} \\
        \addlinespace[2pt]
        Bimanual Box Transfer
        & dual-arm
        & 16
        & \texttt{left\_side}, \texttt{left\_wrist}, \texttt{right\_wrist} \\
        \bottomrule
    \end{tabularx}
\end{table}

\paragraph{Language, vision, and temporal window.}
The policy is vision-language only: proprioception is not provided. The language input is the natural-language instruction of the current task, for example ``pick up the bread'' or ``close the drawer''. When a task has several phrasings, they are stored as a $|$-delimited list and one phrasing is sampled per training example. A text-conditioning dropout of $0.1$ is applied for classifier-free guidance.

Single-arm observations concatenate \texttt{left\_side} and \texttt{left\_wrist} horizontally. The raw mosaic is $224\times 448$ and is resized by the model image pipeline to approximately $192\times 320$. Bimanual observations are packed into a $2\times 2$ mosaic of resolution $256\times 256$, with \texttt{left\_side} and \texttt{right\_wrist} on the top row, \texttt{left\_wrist} on the bottom left, and a black placeholder on the bottom right.

Demonstrations are recorded at $30$\,FPS. Each training example contains 17 video frames and the corresponding 16-step future action chunk. At inference, frame 0 is the visual condition and the model predicts the subsequent 16 video frames together with 16 actions, for a horizon of $16/30\approx 0.53$\,s.

\paragraph{Action representation.}
Actions are absolute joint positions rather than joint deltas. Internally the model uses a padded action tensor of shape $[B,16,64]$; only the first $d_a$ dimensions are valid, and the remaining coordinates are zero-filled and masked in the loss, the injected noise, and the predicted-velocity computation. For single-arm tasks, $d_a=8$ stores seven absolute arm joints (radians) followed by an absolute gripper command. For the bimanual task, $d_a=16$ stores fourteen absolute arm joints followed by the left and right absolute gripper commands. Each dimension is linearly normalized from its 1st/99th-percentile statistics onto $[-1,1]$: the three single-arm tasks share one 8-dimensional absolute-joint statistic, while the bimanual task uses a separate 16-dimensional statistic. Predicted actions are denormalized with the corresponding statistic before execution.

\paragraph{Optimization.}
\Cref{tab:cosmos_edge_hparams} reports the supervised fine-tuning configuration. We train for $40{,}000$ iterations in BF16 (FSDP master weights in FP32) on 8 GPUs with a per-GPU batch size of 32 and no gradient accumulation, giving a global batch size of 256. The optimizer is FusedAdam with a base learning rate of $5\times 10^{-5}$; the action adapters \texttt{action2llm}, \texttt{llm2action}, and \texttt{action\_modality\_embed} use a $5\times$ learning rate of $2.5\times 10^{-4}$. A LambdaLinear schedule warms up for 500 steps over a $40{,}000$-step cycle and decays to a final learning-rate multiplier of 0. Flow matching uses an independent action-time schedule with a logit-normal action-time distribution, so that the action and video branches do not share a flow-matching timestep; a single scalar action $\sigma$ is shared across the 16-step action sequence of each example. Training uses Power EMA with rate $0.1$.

\begin{table}[t]
    \centering
    \caption{\textbf{Cosmos3-Edge four-task SFT hyperparameters.} The recipe fine-tunes a DROID-initialized Cosmos3-Edge world-action model jointly on the four real-world tasks.}
    \label{tab:cosmos_edge_hparams}
    \scriptsize
    \renewcommand{\arraystretch}{1.05}
    \setlength{\tabcolsep}{6pt}
    \begin{tabularx}{\textwidth}{@{}p{0.34\textwidth}X@{}}
        \toprule
        \textbf{Hyperparameter} & \textbf{Value} \\
        \midrule
        Initialization
        & Cosmos3-Edge-Policy-DROID \\
        Reasoner
        & Nemotron-2B \\
        Training algorithm
        & SFT with flow matching \\
        Action representation
        & absolute joint position \\
        Maximum iterations
        & $40{,}000$ \\
        GPUs / per-GPU batch / accumulation
        & $8$ / $32$ / $1$ (global batch $256$) \\
        Trainer / episode-shuffle seed
        & $42$ / $42$ \\
        Precision
        & BF16 activations; FP32 FSDP master weights \\
        Gradient clipping
        & $1.0$ \\
        Checkpoint interval
        & $1{,}000$ iterations \\
        Optimizer
        & FusedAdam with $\beta_1=0.9$, $\beta_2=0.99$, $\epsilon=10^{-8}$, weight decay $0.05$ \\
        Base learning rate
        & $5\times10^{-5}$ \\
        Action-adapter learning rate
        & $2.5\times10^{-4}$ ($5\times$ the base rate) on \texttt{action2llm}, \texttt{llm2action}, \texttt{action\_modality\_embed} \\
        Scheduler
        & LambdaLinear; $500$-step warm-up; $40{,}000$-step cycle; final learning-rate multiplier $0$ \\
        Independent action schedule
        & yes; logit-normal action time; action-time shift $1.0$; uniform train-time weighting \\
        Dynamic shift / discrete rectified flow
        & off / off \\
        Action / vision flow-matching loss weights
        & $10.0$ / $10.0$ \\
        Action $\sigma$
        & one scalar shared across the 16-step action sequence \\
        EMA
        & Power EMA with rate $0.1$ \\
        Text-conditioning dropout
        & $0.1$ \\
        Proprioceptive state
        & not used \\
        Action horizon / padded width
        & $H=16$ / $64$ \\
        Data frame rate
        & $30$\,FPS \\
        \bottomrule
    \end{tabularx}
\end{table}

\paragraph{Trainable parameters.}
The four-task SFT stage is a selected-module fine-tuning procedure rather than a full-parameter update: only the generation tower and the action pathway are trained, while the vision-language understanding tower, the VAE, and all remaining parameters stay frozen. All parameters of the selected modules are updated densely. \Cref{tab:cosmos_edge_trainable} lists the selected modules.

\begin{table}[H]
    \centering
    \caption{\textbf{Selected modules.} The recipe uses selected-module fine-tuning: only the generation tower and the action pathway are optimized.}
    \label{tab:cosmos_edge_trainable}
    \small
    \setlength{\tabcolsep}{6pt}
    \renewcommand{\arraystretch}{1.15}
    \begin{tabularx}{\textwidth}{
        @{}
        >{\raggedright\arraybackslash}p{0.18\textwidth}
        >{\raggedright\arraybackslash}X
        @{}
    }
        \toprule
        \textbf{Status}
        & \textbf{Modules} \\
        \midrule
        Updated
        & generation tower \texttt{moe\_gen}; time embedder \texttt{time\_embedder}; VAE--LLM adapters \texttt{vae2llm}, \texttt{llm2vae}; action pathway \texttt{action2llm}, \texttt{llm2action}, \texttt{action\_modality\_embed} \\
        \addlinespace[2pt]
        Frozen
        & vision-language understanding tower; VAE; all parameters outside the selected keys \\
        \bottomrule
    \end{tabularx}
\end{table}

\subsubsection{Implementation Inputs and Outputs}
\label{app:implementation_io}

The $\pi_{0.5}$ VLA interface is specified in \Cref{app:shared_policy_backbone}. The remaining policy-learning interfaces are the IQL critic and the DSRL actor--critic. Throughout this section, $o_t$ denotes the multi-camera visual observation at policy-decision step $t$, and $a_t$ denotes the corresponding executable action chunk. No component receives a proprioceptive or robot-state vector.

\paragraph{Control and action representation.}
Low-level control runs at $10$\,Hz. During online DSRL, the latent action has shape $H\times d_z$ with $d_z=32$, and the frozen $\pi_{0.5}$ decoder is queried after every $q=10$ executed low-level control steps. The $\pi_{0.5}$ action layout is given in \Cref{app:shared_policy_backbone}.

\paragraph{Camera configuration.}
The dual-arm Franka platform is instrumented with 4 Intel RealSense cameras: 2 D435 third-person cameras, denoted \texttt{left\_side} and \texttt{right\_side}, and 2 D405 wrist-mounted cameras, denoted \texttt{left\_wrist} and \texttt{right\_wrist}. Images are resized with aspect-preserving padding to $224\times224$ for $\pi_{0.5}$ and the IQL critic; the DSRL visual encoder receives inputs downsampled to $64\times64$. \Cref{tab:real_robot_camera_views} lists the exact streams used by each pathway.

\begin{table}[H]
    \centering
    \caption{\textbf{Camera streams used by the policy and RL encoders.} Entries list the exact streams provided to each pathway.}
    \label{tab:real_robot_camera_views}
    \small
    \setlength{\tabcolsep}{6pt}
    \renewcommand{\arraystretch}{1.15}
    \begin{tabularx}{\textwidth}{
        @{}
        >{\raggedright\arraybackslash}p{0.18\textwidth}
        >{\raggedright\arraybackslash}p{0.27\textwidth}
        >{\raggedright\arraybackslash}X
        @{}
    }
        \toprule
        \textbf{Pathway}
        & \textbf{Single-arm tasks}
        & \textbf{Bimanual Box Transfer} \\
        \midrule
        $\pi_{0.5}$
        & \texttt{left\_side}, \texttt{left\_wrist}
        & \texttt{left\_side}, \texttt{left\_wrist}, \texttt{right\_wrist} \\
        \addlinespace[2pt]
        IQL critic
        & \texttt{left\_side}, \texttt{left\_wrist}
        & \texttt{left\_side}, \texttt{left\_wrist}, \texttt{right\_wrist} \\
        \addlinespace[2pt]
        DSRL actor--critic
        & \texttt{left\_side}, \texttt{left\_wrist}
        & \texttt{left\_side}, \texttt{right\_side}, \texttt{left\_wrist}, \texttt{right\_wrist} \\
        \bottomrule
    \end{tabularx}
\end{table}

\paragraph{IQL critic and value.}
The offline RL networks consume the current and next multi-camera observations together with the demonstrated action chunk. They estimate $Q(o,a)$ and $V(o)$, from which we compute the advantage weight used to refine $\pi_{0.5}$.

\paragraph{DSRL actor--critic.}
The online RL networks consume only multi-camera observations at $64\times64$. The actor predicts a latent-noise chunk $z\in[-1,1]^{16\times32}$, the critics estimate its Q-values, and automatic entropy tuning optimizes the temperature $\alpha$.

\paragraph{Reward-model inference.}
RynnValue and Robometer are deployed through the same scoring interface. For each collected trajectory, the reward service receives the task instruction and the RGB sequence from the \texttt{right\_side} third-person camera. Both models operate in a causal, history-conditioned scoring mode. Given a trajectory containing $T$ RGB observations, for each step $t=1,\ldots,T$, the service scores the observation history $o_{1:t}$, uniformly subsampled to 4 frames (rather than the 8 used during training, to match Robometer's 4-frame inference protocol) before inference. Robometer decodes its native normalized progress $p_t\in[0,1]$ as the expectation over ten progress bins and uses $\Phi_t^{\mathrm{Robo}}=p_t$. During reward relabeling and online RL, RynnValue uses only its absolute temporal-distance head and does not invoke the language-generation branch. Following the notation in the main text, we convert its predicted temporal distance $v_t$ into a higher-is-better potential using $\Phi_t=-v_t$. Each complete trajectory is scored once after collection, and the resulting potential sequence is sampled at policy-decision boundaries to construct the shaping rewards. Reward relabeling therefore operates at the action-chunk granularity used by the RL algorithm, rather than at every low-level control step.

\subsubsection{Policy and RL Initialization}
\label{app:rl_initialization}

\paragraph{RL networks.}
The auxiliary RL networks are small relative to the VLA. Offline IQL uses a ResNet-18 visual encoder with GroupNorm, spatial softmax, and a 50-dimensional bottleneck, followed by MLP heads with hidden dimensions $(256,256)$. It maintains 2 Q-functions and a separate value function. Online DSRL uses a 4-layer convolutional visual encoder with 32 channels per layer, GroupNorm, spatial softmax, and a 50-dimensional bottleneck. Its actor and critic MLPs have hidden dimensions $(128,128,128)$; the critic ensemble contains ten Q-functions, and the actor is a $\tanh$-squashed Gaussian distribution over latent noise.

\paragraph{Initialization and training stages.}
SFT and IQL are independent, single-stage fine-tuning procedures initialized from the same pretrained $\pi_{0.5}$ checkpoint; IQL is not initialized from SFT. SFT optimizes the standard flow-matching behavior-cloning objective, whereas IQL uses advantage-weighted flow matching. Online DSRL uses task-specific initialization: Bread Basket Placement and Steak Serving with a Spatula start from their SFT checkpoints, while Box-in-Drawer Placement and Bimanual Box Transfer start from the corresponding Robometer-based offline-RL checkpoints. The initial checkpoint for each task is held fixed across all online reward variants. During DSRL, the initialized VLA remains frozen and serves only as a noise-conditioned action decoder; the latent policy, critic ensemble, and entropy temperature are optimized.

\subsubsection{Offline RL with IQL}
\label{app:offline_iql}

\paragraph{Mixed-expertise dataset construction.}
For each task $j$, we construct a mixed-expertise offline dataset by combining all successful and unsuccessful trajectories:
\begin{equation}
\mathcal{D}_j
=
\mathcal{D}_{j,\mathrm{succ}}
\cup
\mathcal{D}_{j,\mathrm{fail}}.
\end{equation}
Each trajectory is segmented at policy-decision boundaries into transitions
$(o_h,a_h,o_{h+1},m_h)$, where $a_h$ is an action chunk and $m_h\in\{0,1\}$ is the bootstrap mask. We set $m_h=0$ for terminal transitions and $m_h=1$ otherwise.

All reward variants share the same transitions, task-outcome labels, policy initialization, and optimization configuration. We relabel this common dataset with the potential produced by RynnValue or the corresponding baseline reward model; only the potential source and shaping coefficient differ across variants.

\paragraph{Offline reward construction.}
For a transition observation $o_h$, we write $\Phi_h = -v_h$ for its shaping potential, where $v_h$ is RynnValue's decoded absolute temporal-distance prediction for $o_h$ (\Cref{eq:rv_potential}); the instruction $\ell$ and metadata $m$ are omitted for brevity.
The sparse task reward is
\begin{equation}
r_h^{\mathrm{sparse}}
=
\begin{cases}
0,  & \text{if the task is completed after executing $a_h$}, \\
-1, & \text{otherwise}.
\end{cases}
\label{eq:real_world_sparse_reward}
\end{equation}
Thus, every transition before completion receives $-1$, while the transition that completes the task receives $0$. All transitions in an unsuccessful trajectory receive $-1$.

The offline potential-based shaping reward is
\begin{equation}
r_h^{\mathrm{shape}}
=
\gamma_{\mathrm{off}}\Phi_{h+1}
-
\Phi_h,
\end{equation}
where $\gamma_{\mathrm{off}}$ is the offline RL discount factor. The final reward used to train IQL is
\begin{equation}
r_h^{\mathrm{off}}
=
r_h^{\mathrm{sparse}}
+
\kappa
r_h^{\mathrm{shape}},
\label{eq:offline_total_reward}
\end{equation}
where $\kappa\geq0$ is the shaping coefficient used in the main text. We set $\kappa=0.1$ for RynnValue, $\kappa=1.0$ for Robometer, and $\kappa=0$ for the sparse-reward baseline. These values are fixed across tasks.

\paragraph{IQL optimization and VLA refinement.}
We use the standard IQL objective~\cite{kostrikov2021offline}: the value function is trained by expectile regression, each Q-function uses a one-step temporal-difference target, and the 2 target Q-functions are aggregated by their minimum. Target Q-networks are updated by Polyak averaging. Rather than introducing a separate IQL policy, we retain the flow-matching VLA and weight its training loss by
\begin{equation}
w(o,a)
=
\min\left\{
\exp\left(
\beta
\left[
\min_k Q_{\theta_k}(o,a)-V_\psi(o)
\right]
\right),
w_{\max}
\right\}.
\end{equation}
For the first $N_{\mathrm{warm}}$ optimization steps, we set $w(o,a)=1$; thereafter, demonstrated actions with larger estimated advantages receive greater weight. \Cref{tab:iql_hparams} reports the complete architecture and optimization configuration.

\begin{table*}[t]
    \centering
    \caption{\textbf{Offline IQL and SFT hyperparameters.} All IQL reward variants share the same mixed-expertise dataset, sparse task reward, policy initialization, and optimization configuration. SFT uses the same policy optimizer and learning-rate schedule and is also trained for $10{,}000$ steps per task.}
    \label{tab:iql_hparams}
    \scriptsize
    \renewcommand{\arraystretch}{1.05}
    \setlength{\tabcolsep}{6pt}
    \begin{tabularx}{\textwidth}{@{}p{0.31\textwidth}X@{}}
        \toprule
        \textbf{Hyperparameter} & \textbf{Value} \\
        \midrule

        Base policy
        & $\pi_{0.5}$ with flow matching; action dimension $32$ \\

        Action horizon $H$
        & $16$ \\

        Batch size
        & $64$ \\

        Policy optimizer
        & AdamW with $\beta_1=0.9$, $\beta_2=0.95$, $\epsilon=10^{-8}$, weight decay $10^{-10}$, and gradient-norm clipping at $1.0$ \\

        Learning-rate schedule
        & Cosine decay with $2{,}000$ linear warm-up steps \\

        Peak / final policy learning rate
        & $3\times10^{-5}$ / $3\times10^{-6}$ \\

        Policy EMA decay
        & $0.99$ \\

        Training steps per task
        & $10{,}000$ \\

        Critic / value optimizer
        & Adam with a learning rate of $3\times10^{-4}$ \\

        Offline discount $\gamma_{\mathrm{off}}$
        & $0.99$ \\

        Target update rate $\rho_{\mathrm{off}}$
        & $0.005$ \\

        Expectile parameter $\tau_e$
        & $0.8$ \\

        Advantage temperature $\beta$
        & $10.0$ \\

        Maximum advantage weight $w_{\max}$
        & $100$ \\

        Number of Q-functions $K_{\mathrm{IQL}}$
        & $2$; minimum aggregation \\

        Critic and value encoder
        & ResNet-18 with GroupNorm and spatial softmax; 50-dimensional bottleneck \\

        Critic and value hidden dimensions
        & $(256,256)$ \\

        Number of critic cameras
        & $2$ for single-arm tasks and $3$ for the bimanual task \\

        Critic input resolution
        & $224\times224$ using the VLA preprocessing pipeline \\

        Policy warm-up $N_{\mathrm{warm}}$
        & $200$ optimization steps with $w(o,a)=1$ \\

        Sparse task reward
        & $-1$ before task completion and $0$ upon task completion \\

        Potential-based shaping reward
        & $r_h^{\mathrm{shape}}
        =\gamma_{\mathrm{off}}\Phi_{h+1}-\Phi_h$ \\

        Shaping coefficient $\kappa$
        & $0.1$ for RynnValue and $1.0$ for Robometer; fixed across tasks \\

        Sparse-reward baseline
        & $\kappa=0$ \\

        Image augmentation
        & Random cropping applied to both current and next observations; no color jitter \\
        \bottomrule
    \end{tabularx}
\end{table*}

\subsubsection{Online RL with DSRL}
\label{app:online_dsrl}

\paragraph{Latent policy optimization.}
For online RL, we use Diffusion Steering via Reinforcement Learning (DSRL)~\cite{wagenmaker2025steering}. The task-specific VLA checkpoint described above is frozen and used as a noise-conditioned action decoder:
\begin{equation}
a_t=\pi_{0.5}(o_t,z_t).
\end{equation}
A Soft Actor-Critic (SAC) policy $\mu_\varphi(z_t\mid o_t)$ is trained over the latent noise $z_t$, rather than directly over executable robot actions. The latent action space is
\begin{equation}
z_t\in[-1,1]^{H\times d_z},
\end{equation}
where $H=16$ is the action horizon and $d_z=32$ is the per-step latent dimension.

\paragraph{SAC optimization.}
We use the standard entropy-regularized SAC objective in the latent action space. Critic targets use the mean of ten target Q-functions, the actor trades off the ensemble-mean Q-value against policy entropy, and the temperature is optimized automatically toward target entropy $\overline{\mathcal{H}}=-\dim(z)$. Target critics are updated by Polyak averaging. \Cref{tab:dsrl_hparams} reports the complete architecture and optimization configuration.

\paragraph{Online reward construction.}
Online DSRL uses the same sparse task-reward convention as \Cref{eq:real_world_sparse_reward}: each transition receives $-1$ until task completion, and the completing transition receives $0$.

Let $\gamma_s$ denote the per-environment-step shaping discount. Because each policy decision executes $q$ low-level control steps, the corresponding chunk-level discount is
\begin{equation}
\gamma_c=\gamma_s^q,
\end{equation}
and the online shaping reward is
\begin{equation}
r_t^{\mathrm{shape}}
=
\begin{cases}
\gamma_c\Phi_{t+1}-\Phi_t,
& \text{if $o_{t+1}$ is nonterminal}, \\[3pt]
-\Phi_t,
& \text{if $o_{t+1}$ is terminal}.
\end{cases}
\end{equation}
We use the shaping coefficient $\kappa$ defined in the main text, setting $\kappa=0.1$ for RynnValue, $\kappa=1.0$ for Robometer, and $\kappa=0$ for the sparse-reward baseline. These values are fixed across tasks, and all variants otherwise share the same reward definition and online optimization configuration. The final reward used by SAC is
\begin{equation}
r_t^{\mathrm{on}}
=
r_t^{\mathrm{sparse}}
+
\kappa
r_t^{\mathrm{shape}}.
\end{equation}

\begin{table}[H]
    \centering
    \caption{\textbf{Online DSRL hyperparameters.} SAC operates in the latent space of the frozen VLA, which decodes latent variables into executable action chunks. All reward variants share the same optimization configuration and differ only in the potential source and shaping coefficient.}
    \label{tab:dsrl_hparams}
    \scriptsize
    \renewcommand{\arraystretch}{1.05}
    \setlength{\tabcolsep}{6pt}
    \begin{tabularx}{\textwidth}{@{}p{0.31\textwidth}X@{}}
        \toprule
        \textbf{Hyperparameter} & \textbf{Value} \\
        \midrule

        Base policy
        & Frozen SFT checkpoint for Bread Basket Placement and Steak Serving with a Spatula; frozen Robometer offline-RL checkpoint for Box-in-Drawer Placement and Bimanual Box Transfer \\

        Latent action space
        & $z\in[-1,1]^{H\times d_z}$, with $H=16$ and $d_z=32$ \\

        RL algorithm
        & SAC with automatic entropy tuning and initial temperature $\alpha_0=1.0$ \\

        Actor optimizer
        & Adam with a learning rate of $1\times10^{-4}$ \\

        Critic optimizer
        & Adam with a learning rate of $3\times10^{-4}$ \\

        Temperature optimizer
        & Adam with a learning rate of $3\times10^{-4}$ \\

        Gradient clipping
        & None \\

        Target entropy $\overline{\mathcal{H}}$
        & $-\dim(z)$ \\

        Online SAC discount $\gamma_{\mathrm{on}}$
        & $0.999$ \\

        Target update rate $\rho_{\mathrm{on}}$
        & $0.005$ \\

        Number of Q-functions $K_{\mathrm{SAC}}$
        & $10$; mean aggregation \\

        Actor and critic hidden dimensions
        & $(128,128,128)$ \\

        Image encoder
        & Four-layer CNN with $32$ channels per layer, strides $(2,1,1,1)$, VALID padding, GroupNorm, spatial softmax, and a 50-dimensional bottleneck \\

        SAC input resolution
        & $64\times64$ \\

        Batch size
        & $256$ \\

        Update-to-data ratio
        & $100$ \\

        Training length
        & $6{,}000$ training steps \\

        Online rollout trajectories
        & $60$ per task \\

        Replay-buffer capacity
        & $\max(\text{training steps}/\text{UTD},10^4)=10^4$ \\

        Update frequency
        & After each episode \\

        Exploration warm-up
        & $N_{\mathrm{noise}}=2$ episodes with Gaussian noise standard deviation $\sigma=0.1$ \\

        Minimum replay size $N_{\mathrm{start}}$
        & $200$ transitions \\

        Maximum episode length
        & $600$ environment steps \\

        Policy-decision frequency
        & Low-level control at $10$ Hz, with one policy decision every $q=10$ environment steps \\

        Number of DSRL cameras
        & $2$ for single-arm tasks and $4$ for the bimanual task \\

        Sparse task reward
        & $-1$ before task completion and $0$ upon task completion \\

        Shaping coefficient $\kappa$
        & $0.1$ for RynnValue, $1.0$ for Robometer, and $0$ for the sparse-reward baseline; fixed across tasks \\

        Per-step shaping discount $\gamma_s$
        & $0.999$ \\
        \bottomrule
    \end{tabularx}
\end{table}

\subsubsection{Scalable Post-Training with Data Filtering}
\label{app:data_filtering}

This section describes the data-filtering procedure used by the Filtered BC baseline in \Cref{sec:scalable_post_training}. Filtered BC identifies and excludes low-quality action chunks based on task progress, then trains a standard behavior-cloning policy on the retained demonstration frames. Filtering is performed offline at the frame level; subsequent policy training requires neither an additional critic or value function nor policy-side conditioning.

\paragraph{Progress-based filtering strategy.}
Given a demonstration trajectory $\{(o_t, a_t)\}_{t=0}^{T-1}$, RynnValue assigns a scalar task-progress score $p_t$ to each observation. The score reflects task completion and is expected to increase as the task progresses. To identify frames associated with forward progress, we define a windowed advantage using only the resulting score sequence. Specifically, the per-frame reward is the progress increment
\begin{equation}
r_t \;=\; p_{t+1} - p_t , \qquad r_{T-1} \;=\; 0 ,
\end{equation}
where the final-frame reward is set to zero because no subsequent frame is available. The windowed advantage at frame $t$ is the $\lambda$-discounted sum of these increments over a window of length $W$:
\begin{equation}
A_t \;=\; \sum_{k=0}^{W-1} \lambda^{k}\, r_{t+k},
\qquad\text{with } r_{t+k} \equiv 0 \text{ for } t+k \ge T .
\label{eq:filter_bc_windowed_advantage}
\end{equation}
Rewards beyond the episode boundary are set to zero, yielding a truncated return for frames near the end. Advantages are computed independently for each episode, so no window extends into another episode.

The windowed advantage therefore measures near-term progress toward task completion. All reported runs use $\lambda = 0.98$, which places greater weight on progress made earlier in the window. A frame is retained if and only if its windowed advantage is strictly positive:
\begin{equation}
\mathrm{keep}(t) \;\iff\; A_t > 0 .
\end{equation}
This rule retains frames followed by net forward progress within the window and discards frames whose windowed progress return is zero or negative.

\paragraph{Filtering and training hyperparameters.}
The evaluated Filtered BC setup uses $\lambda = 0.98$ and retains exactly those frames with strictly positive windowed progress returns. \Cref{tab:filter_bc_hparams} summarizes its filtering and standard BC optimization settings.

\begin{table}[H]
    \centering
    \caption{\textbf{Hyperparameters.} Filtering and optimization settings for the evaluated Franka setup, which retains frames with $A_t>0$ and uses $\lambda=0.98$.}
    \label{tab:filter_bc_hparams}
    \small
    \renewcommand{\arraystretch}{1.15}
    \setlength{\tabcolsep}{6pt}
    \begin{tabularx}{\textwidth}{@{}l X@{}}
        \toprule
        \textbf{Hyperparameter} & \textbf{Value} \\
        \midrule
        Tasks & 4 (drawer, box, bread, steak) \\
        \addlinespace[2pt]
        Retention criterion & $A_t>0$ \\
        Window $W$ & 50 steps \\
        Discount $\lambda$ & 0.98 \\
        Action horizon $H$ & 16 \\
        Action dimension & 32 \\
        Batch size & 64 \\
        Peak LR / schedule & $3\times10^{-5}$, cosine (2k-step warmup, 20k-step decay $\to 3\times10^{-6}$) \\
        EMA decay & 0.99 \\
        Training steps & 20{,}000 \\
        Initialization & $\pi_{0.5}$ base \\
        \bottomrule
    \end{tabularx}
\end{table}

\emph{Progress-source ablation.} For the Franka setup, we replace RynnValue-derived progress scores with Robometer success scores while keeping all filtering and training hyperparameters fixed. This comparison isolates the effect of the progress signal on data selection and subsequent policy learning.

\subsubsection{Inference-time Value Guidance}
\label{app:value_guidance}

We evaluate value guidance on Bread Basket Placement and Steak Serving with a Spatula (\Cref{sec:value_guided_inference}), using the Cosmos3-Edge policy jointly fine-tuned on all four tasks (\Cref{app:cosmos_edge_policy}). The policy and reward model remain frozen throughout evaluation. At each decision step, we sample several candidate actions and use RynnValue to select the one whose predicted outcome makes the most progress toward the goal.

\paragraph{Value-guided action selection.}
Given the current observation $o_t$ and instruction $\ell$, the policy samples $N=4$ candidate action chunks $\{a_t^{(i)}\}_{i=1}^{N}$ with different noise seeds. Each candidate contains $H=16$ actions and is jointly generated with a video of $T=17$ frames: the current conditioning frame followed by 16 imagined future frames. Since RynnValue takes visual observations rather than actions as input, we evaluate each action chunk through its associated video.

Let $\Phi_j^{(i)}=-v_j^{(i)}$ denote the potential assigned to frame $j$ of candidate $i$, where $v_j^{(i)}$ is the absolute temporal-distance prediction conditioned on $\ell$ (\Cref{eq:rv_potential}). We score a candidate by the increase in potential from the current frame to the final predicted frame:
\begin{equation}
    \widehat{Q}_i = \Phi_{T-1}^{(i)}-\Phi_0^{(i)},
    \qquad
    i^{\star}=\arg\max_{i\in\{1,\ldots,N\}}\widehat{Q}_i.
    \label{eq:value_guidance_selection}
\end{equation}
The score measures the predicted reduction in remaining task time over the action horizon. Taking a difference also cancels additive offsets shared by the current and future frame estimates. We execute the selected chunk $a_t^{(i^{\star})}$ and repeat the procedure at the next policy query. This gives a model-based best-of-$N$ selection rule without training an action-conditioned critic.

\paragraph{Scoring imagined trajectories.}
For these two single-arm tasks, we remove reflection padding after VAE decoding and extract the \texttt{left\_side} third-person view from the left half of the camera mosaic. We score $K=5$ frames per candidate, retaining frame $0$ and uniformly sampling four future frames with rounded indices, which gives $\{0,1,6,11,16\}$. All candidate clips are scored in one batch with the original task instruction. Each clip is processed in a single forward pass, rather than repeatedly evaluating its successive prefixes. For the Robometer comparison, we substitute its native progress predictions for the RynnValue potentials while keeping the policy, candidate budget, and selection rule unchanged.

\paragraph{Inference hyperparameters.}
We use the multi-task SFT checkpoint at $40{,}000$ training steps and the sampling configuration in \Cref{tab:value_guidance_hparams}. Both guided variants use the same configuration; only the model used to score the imagined trajectories differs.

\begin{table}[H]
    \centering
    \caption{\textbf{Value-guidance hyperparameters.} Sampling and action-selection settings for evaluating the frozen multi-task Cosmos3-Edge policy on Bread and Steak.}
    \label{tab:value_guidance_hparams}
    \small
    \renewcommand{\arraystretch}{1.15}
    \setlength{\tabcolsep}{6pt}
    \begin{tabularx}{\textwidth}{@{}l X@{}}
        \toprule
        \textbf{Hyperparameter} & \textbf{Value} \\
        \midrule
        Policy checkpoint & Multi-task SFT, $40{,}000$ steps \\
        Candidate count $N$ & 4 \\
        Action horizon $H$ & 16 \\
        Executable action dimension & 8 (single-arm) \\
        Action representation & Absolute joint and gripper commands \\
        Video length $T$ & 17 (one current and 16 future frames) \\
        Scored frames $K$ & 5, at indices $\{0,1,6,11,16\}$ \\
        Sampler / denoising steps & UniPC / 4 \\
        Classifier-free guidance scale & 3.0 \\
        Sampler shift & 5.0 \\
        Conditioning frame rate & $30$\,FPS \\
        Score aggregation / selection & Potential gain / $\arg\max$ \\
        Candidate seeds (deterministic mode) & $\{0,1,2,3\}$ \\
        Executed steps / control frequency & 16 / $30$\,Hz \\
        \bottomrule
    \end{tabularx}
\end{table}

\paragraph{Inference cost.}
In the single-arm setup with $N=4$ and $K=5$, candidate generation and VAE decoding take approximately $1.10$\,s, and value scoring takes $0.33$\,s, giving a total latency of about $1.44$\,s per policy query. The unguided policy takes approximately $0.15$\,s. Reducing $N$ to 2 lowers guided inference to about $0.8$\,s. A 16-step chunk takes $0.53$\,s to execute at $30$\,Hz, so the synchronous implementation incurs waiting time between chunks despite any reduction in the number of chunks needed to complete a task.

\subsection{Compute Resources}
\label{app:compute_resources}

Offline IQL and SFT each use 2 GPUs with 80GB of memory each. IQL assigns one FSDP device to each replica, whereas SFT uses two-way model sharding. Training one task requires approximately 16 hours for offline IQL and 6 hours for SFT. Online DSRL runs on a single x86 server equipped with two Intel Xeon Platinum 8575C processors (48 physical cores per socket with hyper-threading enabled; 192 logical cores in total), 1.5\,TiB of system memory, and eight NVIDIA GeForce RTX 5090 GPUs with 32\,GB of memory each. The server has a dual-socket NUMA topology and runs NVIDIA driver 570.211.01 with CUDA 12.8. Each task uses $6{,}000$ training steps and $60$ online rollout trajectories, requiring approximately 1.5 hours. All online DSRL experiments use single-node execution.

\end{document}